\documentclass[11pt]{article}

\usepackage[utf8]{inputenc}
\usepackage[T1]{fontenc}
\usepackage{amsmath,amssymb,amsfonts}
\usepackage{graphicx}
\graphicspath{ {./images/} }
\usepackage{booktabs}
\usepackage{multirow}
\usepackage{hyperref}
\usepackage{xcolor}
\usepackage{caption}
\usepackage{subfig}
\usepackage{appendix}
\usepackage[margin=1in]{geometry}
\usepackage{enumitem}
\usepackage{algorithm}
\usepackage{algpseudocode}
\usepackage{array}
\usepackage{makecell}
\usepackage{url}
\usepackage{tcolorbox}
\usepackage{cite}
\usepackage{float}
\usepackage{authblk}

\newcommand{\ours}{VLCE}

\hypersetup{
    colorlinks=true,
    linkcolor=blue,
    citecolor=blue,
    urlcolor=blue
}

\begin{document}

\title{\textbf{VLCE: An External Knowledge Framework for\\Contextual Image Captioning in Disaster Assessment}}

\author[1]{Md. Mahfuzur Rahman\thanks{Corresponding author: mdmahfuzur.rahman@students.cau.edu}}
\author[1]{Kishor Datta Gupta}
\author[2]{Marufa Kamal}
\author[3]{Fahad Rahman}
\author[4]{Sunzida Siddique}
\author[3]{Ahmed Rafi Hasan}
\author[1]{Mohd Ariful Haque}
\author[1]{Roy George}

\affil[1]{Department of Cyber Physical Systems, Clark Atlanta University, Atlanta, GA, USA \\
\texttt{mdmahfuzur.rahman@students.cau.edu}, \texttt{mohdariful.haque@students.cau.edu}, \\
\texttt{kgupta@cau.edu}, \texttt{rgeorge@cau.edu}}
\affil[2]{Department of CSE, BRAC University, Dhaka, Bangladesh \\
\texttt{marufa.kamal1@g.bracu.ac.bd}}
\affil[3]{Department of CSE, United International University, Dhaka, Bangladesh \\
\texttt{frahman203014@bscse.uiu.ac.bd}, \texttt{ahasan191131@bscse.uiu.ac.bd}}
\affil[4]{Department of CSE, Daffodil International University, Dhaka, Bangladesh \\
\texttt{sunzida15-9667@diu.edu.bd}}

\date{}
\maketitle

\begin{abstract}
General-purpose vision-language models (VLMs) such as LLaVA and QwenVL produce descriptions of disaster imagery that lack domain-specific vocabulary and actionable detail. We propose the Vision-Language Caption Enhancer (\ours{}), a framework that integrates external semantic knowledge from ConceptNet and WordNet into the caption generation process for post-disaster satellite and UAV imagery. \ours{} operates in two stages: first, a baseline VLM generates an initial caption conditioned on YOLOv8 object detections; second, a knowledge-enriched sequential model, either a CNN-LSTM or a hierarchical cross-modal Transformer, refines the caption using a vocabulary augmented with 1,566 domain-relevant terms extracted from knowledge graphs. We evaluate \ours{} on two disaster benchmarks: xBD (satellite, 6,369 images, 3 damage classes) and RescueNet (UAV, 4,494 images, 12 damage classes), using CLIPScore for semantic alignment and InfoMetIC for informativeness. On RescueNet with the Transformer decoder, \ours{} with knowledge graph enrichment produces captions preferred over QwenVL baselines in 95.33\% of image pairs on InfoMetIC and 73.64\% on CLIPScore. Qualitative analysis shows that without knowledge graph integration, generated captions exhibit hallucinations, word repetition, and semantic incoherence, whereas knowledge-enriched captions maintain factual consistency and domain-appropriate vocabulary.
\end{abstract}


\section{Introduction}
\label{sec:intro}

Natural disasters cause widespread destruction to infrastructure, ecosystems, and communities. Rapid and accurate assessment of post-disaster damage is critical for coordinating emergency response, allocating resources, and planning recovery efforts. Satellite and unmanned aerial vehicle (UAV) images are now the main sources of data for disaster assessment. They give clear pictures of large areas of land.

Recent advances in vision-language models (VLMs) have enabled the automatic generation of textual descriptions from images. Models such as LLaVA~\cite{liu2023visual} and QwenVL~\cite{bai2023qwenvl} achieve strong performance on general image captioning benchmarks. However, when applied to post-disaster imagery, these models produce generic descriptions, lack domain-specific terminology, and fail to capture the nuanced damage indicators that responders require. A caption stating ``buildings in a residential area'' provides little actionable information compared to one identifying ``severely damaged roof structures with debris fields and evidence of flooding in surrounding roadways.''

\begin{figure}[tbp]
    \centering
    \includegraphics[width=0.5\linewidth]{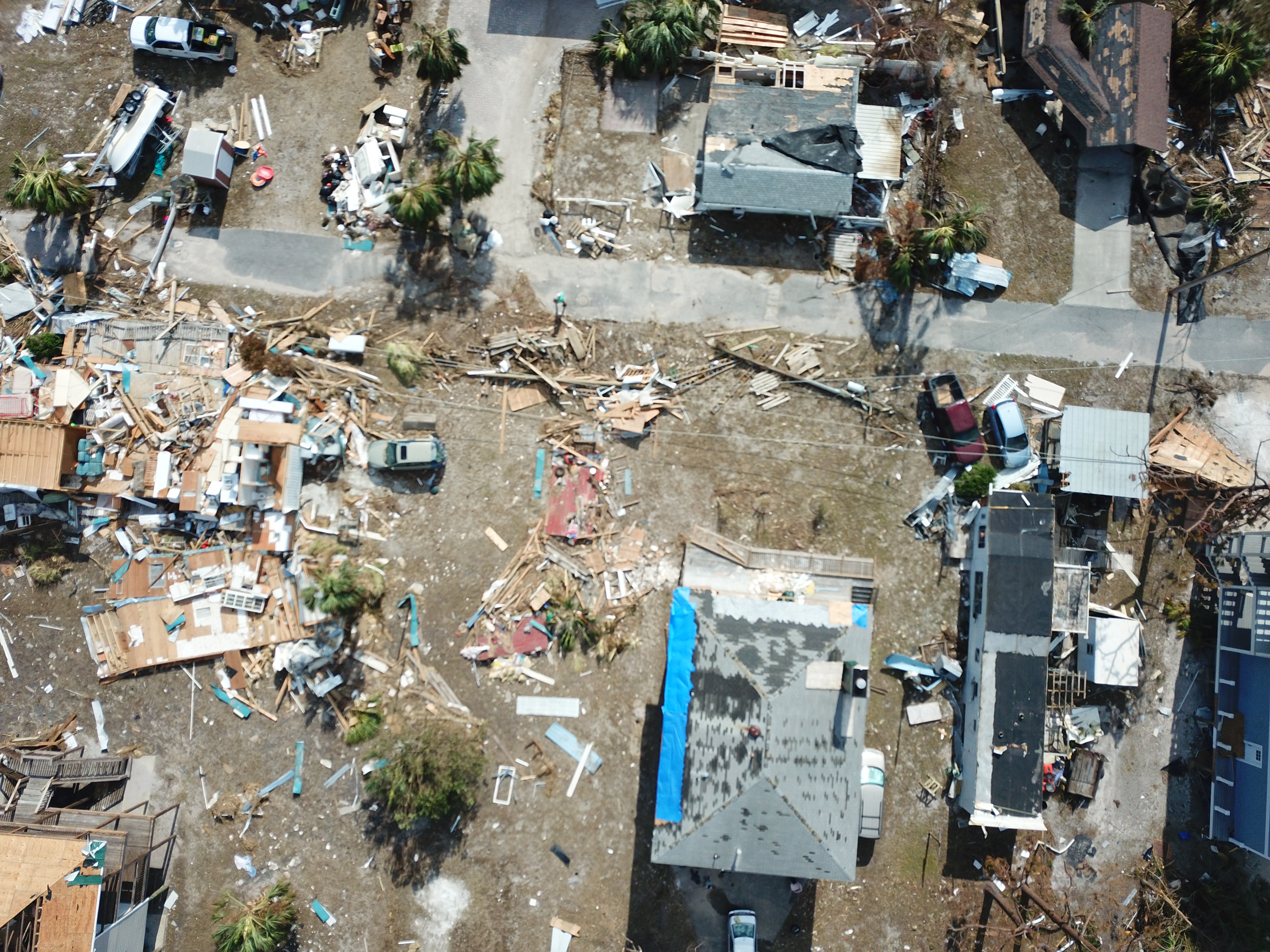}
    \caption{Post-hurricane disaster imagery from the RescueNet dataset.}
    \label{fig:disaster_example}
\end{figure}

This gap arises because general-purpose VLMs are trained mainly on everyday imagery and lack exposure to the specialized vocabulary and visual patterns characteristic of disaster scenarios~\cite{mason2014domain}. The value of contextual captioning is illustrated in Figure~\ref{fig:disaster_example}. A model without domain knowledge produces: \textit{``A satellite image depicts a community from an aerial perspective, revealing multiple dwellings and streets''} In contrast, a knowledge enriched model generates: \textit{``The image shows the aftermath of Hurricane Michael, which inflicted major damage to infrastructure and the environment. Roads and streets are cluttered with debris from destroyed structures, such as fallen trees and scattered items. Trees in the region have suffered varied degrees of damage, with some seeming damaged or totally fallen. Buildings show different degrees of destruction, ranging from comparatively intact ones to those with obvious traces of catastrophic damage. The entire scene depicts the ongoing recovery and rebuilding activities in the aftermath of the catastrophe''}, defining the type of disaster, outlining the effects on infrastructure, and communicating the state of recovery. Adding domain-relevant outside knowledge to the caption generation pipeline is necessary to overcome this constraint.

\subsection{Our Contributions}

We propose \ours{} (Vision-Language Caption Enhancer), a framework that enriches disaster image captioning through external knowledge graph integration. Our contributions are as follows:

\begin{enumerate}[leftmargin=*,itemsep=2pt]
    \item \textbf{Knowledge-enriched captioning framework.} We introduce a two-stage pipeline that combines baseline VLM captions with knowledge graph-augmented sequential models. The system extracts keywords from training captions using RAKE, looks up related terms in ConceptNet and WordNet, and builds a vocabulary of 3,195 words (1,566 from knowledge graphs) that helps create more accurate descriptions of disasters.

    \item \textbf{Dual architecture for satellite and UAV imagery.} We create two different decoder designs: one is a CNN-LSTM model that combines visual and text information, and the other is a hierarchical cross-modal Transformer that processes visual information. Each architecture is paired with a domain-specific image encoder: ResNet50-EuroSAT for satellite imagery and a ViT-based UAV classifier for drone imagery.

    \item \textbf{Comprehensive evaluation on disaster benchmarks.} We evaluate xBD and RescueNet using CLIPScore and InfoMetIC, which are two complementary metrics that together capture both semantic alignment and informational content. We show that knowledge graph integration is essential, and without it, the quality of captions for UAV images drops significantly (less than 2\% preference on both metrics for RescueNet), but with it, the Transformer model gets up to 95.33\% preference compared to standard VLMs on InfoMetIC.
\end{enumerate}

\section{Related Work}
\label{sec:related}

Satellite-based damage assessment has progressed from manual interpretation to automated deep learning pipelines. The xBD dataset~\cite{gupta2019xbd} established a large-scale benchmark for building damage classification from satellite imagery, while RescueNet~\cite{rahnemoonfar2023rescuenet} extended coverage to UAV-captured scenes with fine-grained damage categories. Alisjahbana et al.~\cite{alisjahbana2024deepdamagenet} proposed a two-step CNN for building damage segmentation (0.66 F1). Abbas and Dang~\cite{abbas2023using} investigated CNN-based caption generation for disaster areas (BLEU-1: 0.8731, CIDEr: 5.0908), and Chun et al.~\cite{chun2022deep} applied deep learning for bridge damage descriptions (92.9\% accuracy). Multimodal approaches combining image and text data have shown promise for disaster classification~\cite{zou2021disaster,kota2022multimodal}. Most existing methods focus on damage \emph{classification} rather than free-form description generation.

LLaVA~\cite{liu2023visual} pairs a CLIP ViT-L/14 encoder with a LLaMA decoder through visual instruction tuning. QwenVL~\cite{bai2023qwenvl} extends the Qwen language model with a vision encoder for multi-image understanding. Cornia et al.~\cite{cornia2020meshed} explored meshed-memory Transformers for captioning (CIDEr 132.7 on MS COCO). Chen and Li~\cite{chen2025multi} introduced multi-modal graph aggregation Transformers. While these models excel on general benchmarks, their application to specialized domains such as disaster assessment remains limited by training data distribution~\cite{mason2014domain}.

External knowledge bases have been used to enhance language understanding across a range of tasks. ConceptNet~\cite{schon2019corg} provides commonsense relational knowledge, while WordNet offers lexical relationships including synonymy and hypernymy. ConceptNet Numberbatch embeddings combine distributional semantics with structured knowledge, producing word vectors that encode both corpus statistics and relational information. Zhou et al.~\cite{zhou2019improving} applied knowledge graphs to captioning on MS COCO and Visual Genome. Zhao and Wu~\cite{zhao2023boosting} proposed multi-modal KG methods for graph attention captioning. Tang et al.~\cite{tang2022image} introduced KG-guided attention for CNN-LSTM models. Wang et al.~\cite{wang2024knowledge} explored KG strategies for remote sensing captioning. Despite these advances, KG application to disaster-specific imagery has not been explored.

Pre-trained word embeddings serve as the bridge between discrete vocabulary tokens and continuous neural network inputs. Comparative analyses~\cite{toshevska2020comparative,elbedwehy2023enhanced} have evaluated embeddings for captioning. CLIPScore~\cite{hessel2021clipscore} provides reference-free semantic alignment, while InfoMetIC~\cite{hu2023infometic} assesses informativeness. YOLOv8~\cite{talib2024yolov8} has been adopted for object detection in UAV~\cite{zhai2023yolodrone,wang2023uavyolov8} and general scenarios~\cite{rasheed2025optimized}.

\section{Methodology}
\label{sec:method}

\begin{figure}[ht]
  \centering
  \includegraphics[width=\linewidth]{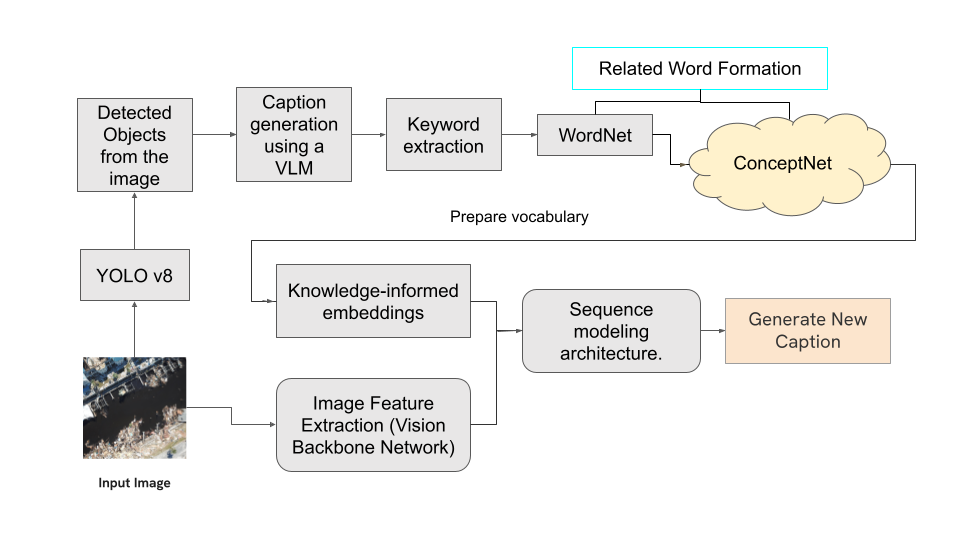}
  \caption{The \ours{} pipeline. YOLOv8 generates scene-aware prompts for a baseline VLM (LLaVA or QwenVL) based on object detection in the input image. ConceptNet and WordNet queries are used to enhance keywords that were taken from training captions using RAKE, creating a vocabulary unique to the domain. The final caption is produced by training a sequential model with the enriched vocabulary and matching embeddings.}
  \label{fig:kg_caption_pipeline}
\end{figure}

Figure~\ref{fig:kg_caption_pipeline} illustrates the \ours{} pipeline. The framework works in two steps: 1) first, it creates basic captions using a ready-made VLM, and 2) it improves those captions with extra knowledge using a trained sequential model.

\subsection{Baseline Caption Generation}
\label{sec:baseline}

We generate initial captions using two VLMs with complementary characteristics. YOLOv8~\cite{talib2024yolov8} first performs object detection on each input image $I$, producing an annotation set $\mathcal{A} = \{(b_i, \ell_i)\}_{i=1}^K$ where the bounding box coordinates $b_i \in \mathbb{R}^4$ and the class label $\ell_i \in \mathcal{L}$. These annotations are integrated into a textual prompt $P(\mathcal{A})$ describing the scene to guide the VLM.

\paragraph{LLaVA-7B v1.5.} We use the LLaVA-7B v1.5 model~\cite{liu2023visual}, made of a CLIP ViT-L/14 visual encoder with a LLaMA language decoder. The encoder produces patch embeddings $V = \phi(I) = [v_1, \ldots, v_N] \in \mathbb{R}^{N \times D}$, concatenated with the prompt into sequence $s$. The decoder generates the caption autoregressively: $\hat{y} = \arg\max_{y} P(y \mid s)$. Generation uses $T = 0.2$ with a maximum output length of 4,000 tokens.

\paragraph{QwenVL-7B Instruct.} We use the QwenVL-7B Instruct model~\cite{bai2023qwenvl} with greedy decoding and a maximum output length of 128 tokens. The vision encoder creates $V = \text{QwenVL}_{\text{vision}}(I) \in \mathbb{R}^{H \times W \times D}$. An event type $e$ and annotations are included in a structured prompt $P(\mathcal{A}, e)$: $\hat{y} = \arg\max_{y} P(y \mid V, P(\mathcal{A}, e))$.

These two baselines differ substantially in output characteristics. LLaVA produces longer, more detailed descriptions, while QwenVL generates concise captions.

\subsection{Knowledge Graph Vocabulary Enrichment}
\label{sec:kg}

The vocabulary enrichment process constructs a domain-specific lexicon from the training captions augmented with external knowledge:

\paragraph{Step 1: Keyword extraction.} We apply the RAKE algorithm to all training captions. With it we extract ranked keyword phrases that capture disaster-relevant concepts such as ``debris field,'' ``structural damage,'' and ``emergency response''.

\paragraph{Step 2: ConceptNet expansion.} For each extracted keyword, we query the ConceptNet API, retrieving up to 10 semantically related terms per keyword. Relation filtering omits synonyms to focus on conceptual connections (e.g., ``hurricane'' $\rightarrow$ ``wind,'' ``flooding,'' ``evacuation'').

\paragraph{Step 3: WordNet synonym enrichment.} Each keyword is additionally queried against WordNet to retrieve synonyms, providing lexical variety. Overlapping substrings are removed to prevent redundancy.

\paragraph{Step 4: Vocabulary construction.} We filter out invalid English words and remove duplicates from the union of the original caption words, ConceptNet terms, and WordNet synonyms. This gives us a final vocabulary of 3,195 tokens, of which 1,566 (49\%) are terms that were not in the original training captions but were added to the knowledge graph. Boundary tokens (\texttt{startseq}, \texttt{endseq}) are added, and sequences are padded to max length 192.

\subsection{Embedding Strategies}
\label{sec:embeddings}

We evaluate two embedding strategies that differ in whether they encode structured knowledge:

\paragraph{With knowledge graph (KG).} ConceptNet Numberbatch embeddings are 300-dimensional word vectors. They are pre-trained to encode both distributional statistics and relational knowledge from ConceptNet. Each term $t$ is mapped to:
\[
\mathbf{e}_t = 
\begin{cases} 
\text{find}(D, t) & \text{if } t \in D \\
\mathbf{e}_{\text{rand}} \sim \mathcal{U}(-\epsilon, \epsilon) & \text{otherwise}
\end{cases}
\]
The embedding layer is initialized with these vectors. It remains frozen during training, preserving structured semantic relationships. The final embedding matrix is $E \in \mathbb{R}^{|V| \times 300}$.

\paragraph{Without KG.} DistilBERT embeddings (768-dimensional) provide contextual representations learned from large-scale text corpora without explicit knowledge graph structure: $e_w = \text{BERT}(w)_{[\text{CLS}]}$. This serves as an ablation to isolate the contribution of structured knowledge.

\subsection{Image Encoders}
\label{sec:encoders}

We select image encoders pre-trained on data distributions matching our target domains:

\paragraph{ResNet50-EuroSAT.} For satellite imagery (xBD), we use a ResNet50 model pre-trained on the EuroSAT land-use classification dataset\footnote{\url{https://huggingface.co/cm93/resnet50-eurosat}}, producing 2,048-dimensional feature vectors. Images are resized to $336 \times 336$ pixels.

\paragraph{ViT-UAV.} For UAV imagery (RescueNet), we use a Vision Transformer pre-trained on a UAV image classification task\footnote{\url{https://huggingface.co/SeyedAli/Remote-Sensing-UAV-image-classification}}, producing 768-dimensional feature vectors via global average pooling. Images are resized to $224 \times 224$ pixels.

\subsection{Sequential Decoders}
\label{sec:decoders}

We design two decoder architectures to investigate how different sequence modeling approaches interact with knowledge-enriched vocabularies.

\subsubsection{CNN-LSTM Decoder}
\label{sec:lstm}

\begin{figure}[tp]
  \centering
  \includegraphics[width=0.3\textwidth]{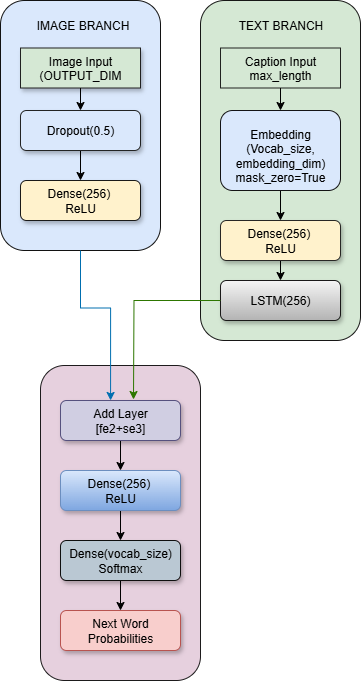}
  \caption{CNN-LSTM architecture with visual and textual branches.}
  \label{fig:lstm_custom}
\end{figure}

The CNN-LSTM decoder (Figure~\ref{fig:lstm_custom}) uses additive fusion of visual and textual features. Given an image feature vector $\mathbf{v} \in \mathbb{R}^{d_v}$ (where $d_v = 2048$ for ResNet50 or $d_v = 768$ for ViT-UAV) and caption tokens $w_1, \ldots, w_T$:

\begin{align}
    \mathbf{f}_{\text{img}} &= \text{ReLU}(\mathbf{W}_v \cdot \text{Dropout}_{0.5}(\mathbf{v})) \in \mathbb{R}^{256} \\
    \mathbf{f}_{\text{txt}} &= \text{LSTM}_{256}(\text{Dropout}_{0.5}(\mathbf{E}[w_1, \ldots, w_T])) \in \mathbb{R}^{256} \\
    \mathbf{h} &= \text{ReLU}(\mathbf{W}_h \cdot (\mathbf{f}_{\text{img}} + \mathbf{f}_{\text{txt}})) \in \mathbb{R}^{256} \\
    P(w_{t+1} | w_{1:t}, \mathbf{v}) &= \text{softmax}(\mathbf{W}_o \cdot \mathbf{h})
\end{align}

\noindent where $\mathbf{E}$ is the embedding matrix (initialized from ConceptNet Numberbatch or DistilBERT and frozen), and $\mathbf{W}_v, \mathbf{W}_h, \mathbf{W}_o$ are learned projection matrices.

\subsubsection{Hierarchical Cross-Modal Transformer Decoder}
\label{sec:transformer}

\begin{figure}[tp]
  \centering
  \includegraphics[width=0.4\textwidth]{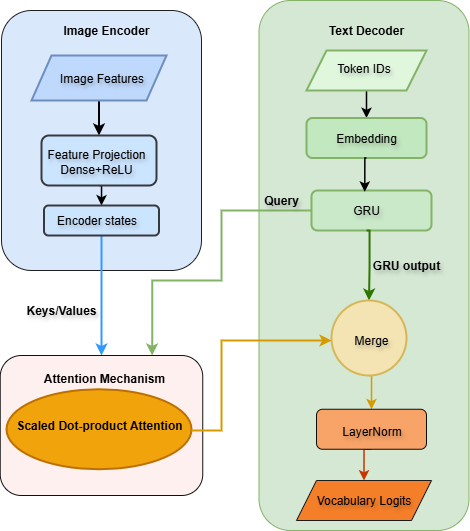}
  \caption{Hierarchical cross-modal Transformer architecture.}
  \label{fig:attention_custom}
\end{figure}

The Transformer decoder (Figure~\ref{fig:attention_custom}) introduces a hierarchical visual encoding scheme that captures multi-scale spatial information, followed by cross-modal attention for visually grounded text generation.

\paragraph{Multi-scale visual encoding.} The image feature vector $\mathbf{v} \in \mathbb{R}^{d_v}$ is projected into three complementary representations:
\begin{align}
    \mathbf{f}_{\text{global}} &= \text{ReLU}(\mathbf{W}_g \mathbf{v} + \mathbf{b}_g) \in \mathbb{R}^{d_{\text{model}}} \\
    \mathbf{F}_{\text{regional}} &= \text{Dense}_{d_{\text{model}}}(\text{Reshape}_{4 \times d_{\text{model}}/4}(\text{ReLU}(\mathbf{W}_r \mathbf{v} + \mathbf{b}_r))) \in \mathbb{R}^{4 \times d_{\text{model}}} \\
    \mathbf{F}_{\text{local}} &= \text{Dense}_{d_{\text{model}}}(\text{Reshape}_{12 \times d_{\text{model}}/12}(\text{ReLU}(\mathbf{W}_l \mathbf{v} + \mathbf{b}_l))) \in \mathbb{R}^{12 \times d_{\text{model}}}
\end{align}

These are concatenated and normalized to form the visual context:
\begin{equation}
    \mathbf{F}_v = \text{LayerNorm}([\mathbf{f}_{\text{global}}; \mathbf{F}_{\text{regional}}; \mathbf{F}_{\text{local}}]) \in \mathbb{R}^{17 \times d_{\text{model}}}
\end{equation}

\paragraph{Cross-modal decoding.} Input tokens are embedded with frozen pre-trained matrix $\mathbf{E}$ and sinusoidal positional encodings: $\mathbf{H}^{(0)} = \mathbf{E}[\mathbf{w}_{1:T}] + \text{PE}(\mathbf{w}_{1:T})$. Each of the $L$ decoder layers applies causal self-attention over caption tokens, followed by cross-attention over the visual context:
\begin{align}
    \mathbf{A}_{\text{self}} &= \text{MultiHead}(\mathbf{Q}, \mathbf{K}, \mathbf{V}, \mathbf{M}_{\text{causal}}) \\
    \mathbf{A}_{\text{cross}} &= \text{MultiHead}(\mathbf{H}, \mathbf{F}_v, \mathbf{F}_v)
\end{align}
where $\mathbf{H}$ is the output of the self-attention sublayer and $\mathbf{M}_{\text{causal}}$ enforces the autoregressive property.

\paragraph{Output generation.} Global visual semantics are fused with textual representations:
\begin{align}
\mathbf{c}_{\text{visual}} &= \operatorname{GlobalAvgPool}(\mathbf{F}_v) \\
\hat{\mathbf{y}}_t &= \operatorname{Dense}_V\Big(\operatorname{LayerNorm}\big([\mathbf{h}_t^{(L)}; \mathbf{c}_{\text{visual}}]\big)\Big)
\end{align}

\subsection{Training}
\label{sec:training}

All models use $d_{\text{model}} = d_{\text{emb}} = 300$, $L = 2$ Transformer layers with $h = 6$ attention heads, and are trained with masked sparse categorical cross-entropy loss:
\begin{equation}
\mathcal{L} = -\frac{1}{\sum_{t=1}^{T} m_t} \sum_{t=1}^{T} m_t \log P(w_t | \mathbf{w}_{<t}, \mathbf{F}_v; \theta)
\end{equation}
where $m_t \in \{0,1\}$ masks padding positions. Batch size is 32, maximum sequence length 192 tokens. Captions are bounded by \texttt{startseq} and \texttt{endseq} tokens. At inference, generation proceeds autoregressively until the stop token or maximum length is reached. The complete pipeline is formalized in Algorithm~\ref{alg1}.

\begin{algorithm}[ht]
\caption{\ours{}: Knowledge Graph-Enhanced Caption Refinement}
\label{alg1}
\begin{algorithmic}[1]
\Require Disaster image dataset $\mathcal{D} = \{I_1, \dots, I_n\}$
\Ensure Enhanced captions $\mathcal{C}_{\text{Enhanced}}$
\Statex
\State \textbf{Stage 1: Baseline Caption Generation}
\For{each image $I_i \in \mathcal{D}$}
    \State $O_i \leftarrow \text{YOLOv8}(I_i)$; $P_i \leftarrow \text{ConstructPrompt}(I_i, O_i)$
    \State $C_i^{\text{VLM}} \leftarrow \text{VLM}(I_i, P_i)$
\EndFor
\Statex
\State \textbf{Stage 2: Knowledge Preparation}
\State $\mathcal{V}_{\text{Enriched}} \leftarrow \text{Vocab}(\{C_i^{\text{VLM}}\}) \cup \text{RAKE+ConceptNet+WordNet}$
\State $E \leftarrow \text{BuildEmbeddings}(\mathcal{V}_{\text{Enriched}})$
\Statex
\State \textbf{Stage 3: Training}
\State $\mathcal{F}_{\text{Img}} \leftarrow E_{\text{Visual}}(\mathcal{D})$; Train $\mathcal{N}_\theta$ on $(\mathcal{F}_{\text{Img}}, \mathcal{V}_{\text{Enriched}}, E)$
\Statex
\State \textbf{Stage 4: Enhanced Caption Generation}
\For{each image $I_i$}
    \State $C_i^{\text{Enhanced}} \leftarrow \text{Decode}(I_i, \mathcal{N}_\theta)$
\EndFor
\State \textbf{return} $\mathcal{C}_{\text{Enhanced}}$
\end{algorithmic}
\end{algorithm}

\section{Experimental Setup}
\label{sec:experiments}

\subsection{Datasets}
\label{sec:datasets}

\paragraph{xBD~\cite{gupta2019xbd}.} A satellite imagery dataset for building damage assessment containing 12,738 images. We use 6,369 post-disaster images, merging ``major-damage'' and ``minor-damage'' into a single ``damaged'' class to yield three labels: no-damage, damaged, and destroyed. Split: 80/20 (5,095 train / 1,274 test).

\paragraph{RescueNet~\cite{rahnemoonfar2023rescuenet}.} A UAV imagery dataset captured after Hurricane Michael, containing 4,494 images at $3{,}000 \times 4{,}000$ pixel resolution across 12 fine-grained damage categories. Split: 80/20.

\subsection{Evaluation Metrics}
\label{sec:metrics}

We evaluate using two complementary metrics:

\paragraph{CLIPScore~\cite{hessel2021clipscore}.} Computes the cosine similarity between CLIP embeddings (\texttt{openai/\hspace{0pt}clip-\hspace{0pt}vit-\hspace{0pt}base-\hspace{0pt}patch32}) of the image and generated caption. For each test image, we compare the CLIPScore of the \ours{}-generated caption against the baseline VLM caption and note the percentage of images where \ours{} scores higher.

\paragraph{InfoMetIC~\cite{hu2023infometic}.} A composite metric: $\alpha \cdot \text{Informativeness} + \beta \cdot \text{Relevance} + \gamma \cdot \text{Precision}$, where \text{Informativeness} is computed as the average negative log-probability of caption words conditioned on the image using a vision-language model. InfoMetIC penalizes generic descriptions and rewards information-rich captions.

\subsection{Configurations}

We associate each decoder architecture with a designated baseline VLM: CNN-LSTM with LLaVA and Transformer with QwenVL. Each pairing is assessed with and without knowledge graph enhancement across both datasets (RescueNet/ViT-UAV and xBD/ResNet50-EuroSAT).

\begin{table}[ht]
\centering
\caption{Percentage of test images where \ours{} captions score higher than the baseline VLM caption. CNN-LSTM models are compared against LLaVA; Transformer models against QwenVL. Bold indicates \ours{} outperforms the baseline on the majority of images.}
\label{tab:main_results}
\small
\renewcommand{\arraystretch}{1.15}
\begin{tabular}{@{}llll cc cc@{}}
\toprule
& & & & \multicolumn{2}{c}{\textbf{CLIPScore (\%)}} & \multicolumn{2}{c}{\textbf{InfoMetIC (\%)}} \\
\cmidrule(lr){5-6} \cmidrule(lr){7-8}
\textbf{Decoder} & \textbf{Baseline} & \textbf{Dataset} & \textbf{KG} & \ours{} & Baseline & \ours{} & Baseline \\
\midrule
\multirow{4}{*}{CNN-LSTM}
& \multirow{4}{*}{LLaVA}
& RescueNet & Yes & \textbf{52.95} & 47.05 & \textbf{54.51} & 45.49 \\
& & RescueNet & No & 0.56 & 99.44 & 1.22 & 98.78 \\
& & xBD & Yes & \textbf{51.10} & 48.90 & \textbf{66.56} & 33.44 \\
& & xBD & No & \textbf{55.34} & 44.66 & \textbf{66.41} & 33.59 \\
\midrule
\multirow{4}{*}{Transformer}
& \multirow{4}{*}{QwenVL}
& RescueNet & Yes & \textbf{73.64} & 26.36 & \textbf{95.33} & 4.67 \\
& & RescueNet & No & 0.22 & 99.78 & 0.08 & 99.92 \\
& & xBD & Yes & \textbf{60.60} & 39.40 & \textbf{69.86} & 30.14 \\
& & xBD & No & 22.14 & 77.86 & 18.76 & 81.24 \\
\bottomrule
\end{tabular}
\end{table}

\begin{table}[ht]
\centering
\caption{Relevant object counts across model configurations.}
\label{tab:noun_analysis}
\small
\begin{tabular}{@{}clcccc@{}}
\toprule
\textbf{Exp.} & \textbf{Configuration} & \textbf{\ours{}} & \textbf{LLaVA} & \textbf{QwenVL} & \textbf{Best} \\ 
\midrule
\multicolumn{6}{c}{\textbf{ViT-UAV / RescueNet}} \\ 
\midrule
1 & With Knowledge Graph      & \textbf{272} & 185 & 201 & \ours{} \\
2 & Without Knowledge Graph   & \textbf{272} & 178 & 195 & \ours{} \\
\midrule
\multicolumn{6}{c}{\textbf{ResNet-EuroSAT / xBD}} \\ 
\midrule
3 & With Knowledge Graph      & \textbf{640} & 445 & 463 & \ours{} \\
4 & Without Knowledge Graph   & \textbf{640} & 431 & 448 & \ours{} \\
\midrule
\multicolumn{6}{c}{\textbf{Summary}} \\ 
\midrule
\multicolumn{2}{c}{Avg.\ improvement over LLaVA} & \textbf{+45.2\%} & -- & -- & -- \\
\multicolumn{2}{c}{Avg.\ improvement over QwenVL} & \textbf{+38.7\%} & -- & -- & -- \\
\multicolumn{2}{c}{Experiments won} & \textbf{4/4} & 0/4 & 0/4 & -- \\
\bottomrule
\end{tabular}
\end{table}


\section{Results}
\label{sec:results}

\subsection{Main Results}

Table~\ref{tab:main_results} presents the percentage of test images where \ours{} captions score higher than the corresponding baseline VLM caption. Values above 50\% indicate \ours{} outperforms the baseline for the majority of images.

\subsection{Key Findings}

\begin{figure*}[hp]
  \centering

  \begin{minipage}{\textwidth}
    \centering
    \includegraphics[width=0.7\textwidth]{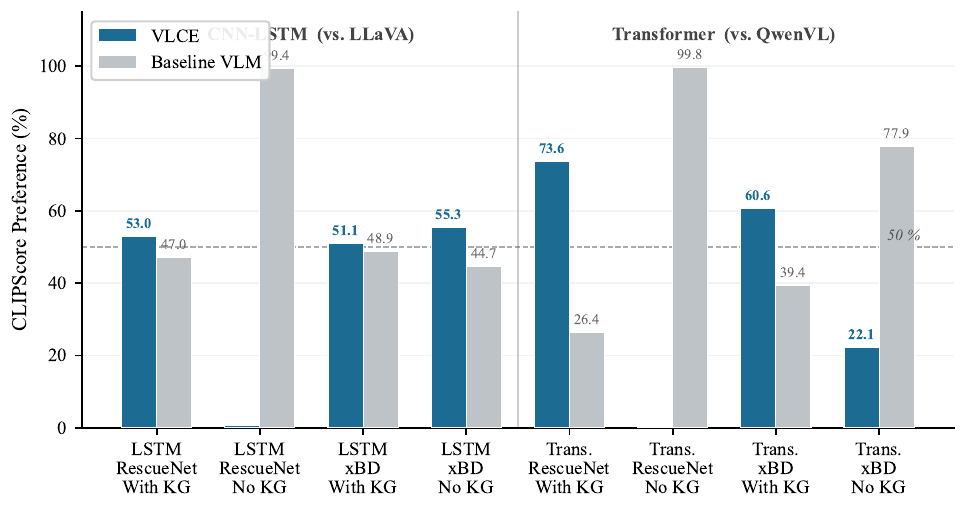}
    \captionof{figure}{CLIPScore preference across all configurations. The dashed line marks 50\% (parity). Without KG, \ours{} preference collapses on RescueNet.}
    \label{fig:clipscore_pref}
  \end{minipage}

  \vspace{6pt}

  \begin{minipage}{\textwidth}
    \centering
    \includegraphics[width=0.7\textwidth]{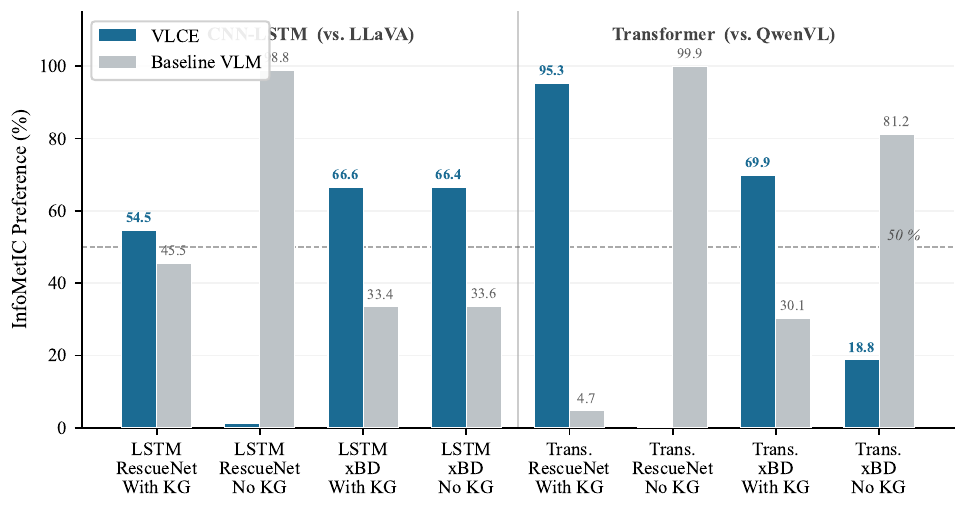}
    \captionof{figure}{InfoMetIC preference across all configurations. The Transformer with KG on RescueNet achieves 95.33\% preference over QwenVL.}
    \label{fig:infometic_pref}
  \end{minipage}

  \vspace{6pt}

  \begin{minipage}{0.48\textwidth}
    \centering
    \includegraphics[width=\textwidth]{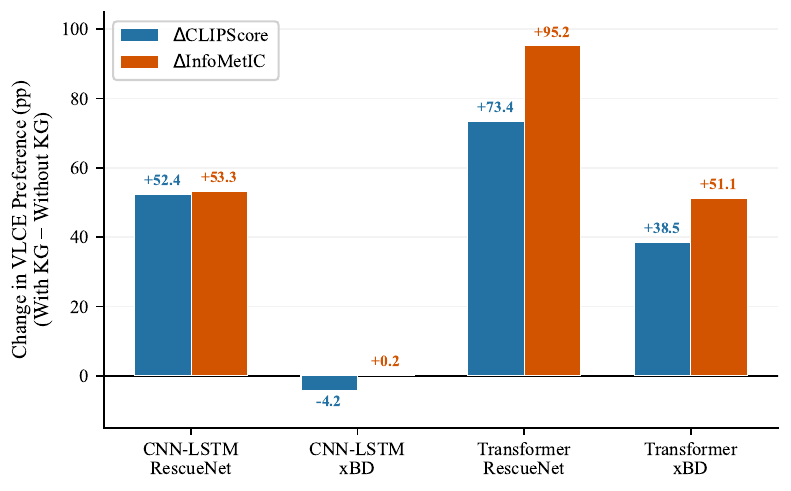}
    \captionof{figure}{Effect of KG integration measured as the change in \ours{} preference (pp). The Transformer on RescueNet gains +73.4 pp CLIPScore and +95.3 pp InfoMetIC.}
    \label{fig:kg_ablation}
  \end{minipage}
  \hfill
  \begin{minipage}{0.48\textwidth}
    \centering
    \includegraphics[width=\textwidth]{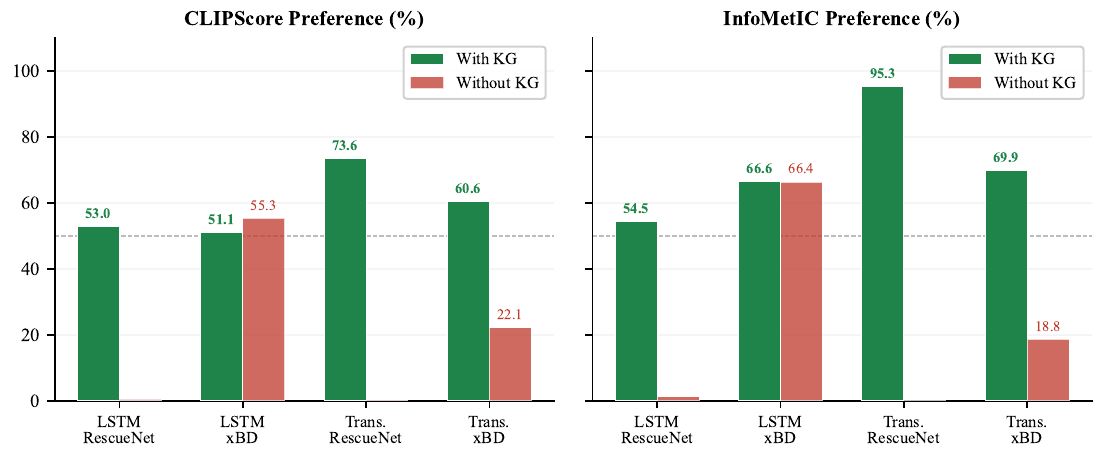}
    \captionof{figure}{With vs.\ without KG. Green bars consistently exceed 50\%, while red bars collapse to near-zero on RescueNet.}
    \label{fig:kg_comparison}
  \end{minipage}

\end{figure*}

\paragraph{Knowledge graph integration is critical for UAV imagery.}
On RescueNet, removal of knowledge graph (KG) enrichment causes near-complete performance collapse for both architectures (Figure~\ref{fig:kg_comparison}). The CNN-LSTM without KG achieves only 0.56\% CLIPScore preference and 1.22\% InfoMetIC preference. The Transformer without KG drops to 0.22\% and 0.08\%, respectively. The 12-class damage taxonomy of RescueNet requires exact terminology that the model cannot learn from limited training captions alone.

\paragraph{The Transformer with KG achieves the strongest results.}
The best configuration is the Transformer decoder with ViT-UAV encoder and KG enrichment on RescueNet, achieving 73.64\% CLIPScore and 95.33\% InfoMetIC preference over QwenVL (Figure~\ref{fig:clipscore_pref}, Figure~\ref{fig:infometic_pref}). The gap between metrics indicates \ours{} captions are substantially more \emph{informative} than the QwenVL baseline, even in cases where visual alignment scores are competitive.

\paragraph{Satellite imagery is more forgiving.}
On xBD, CNN-LSTM without KG still achieves 55.34\% CLIPScore and 66.41\% InfoMetIC preference. The lower label complexity (3 vs.\ 12 classes) and the ResNet50-EuroSAT encoder's satellite-specific pre-training partially compensate for the absence of knowledge-enriched vocabulary.

\paragraph{Architecture-specific patterns.}
The CNN-LSTM shows binary behavior on RescueNet: it works with KG or fails without. The Transformer exhibits more graceful degradation on xBD without KG. It achieves a 22.14\% CLIPScore, suggesting cross-attention can partially compensate for missing vocabulary knowledge on simpler scenes.

\subsection{Quantitative Analysis}

Figures~\ref{fig:clipscore_pref}--\ref{fig:kg_comparison} aggregate the quantitative findings. Two observations are noteworthy. The KG ablation (Figure~\ref{fig:kg_ablation}) demonstrates that the Transformer architecture derives much bigger advantages from KG on RescueNet (+73.4 pp CLIPScore, +95.3 pp InfoMetIC) compared to xBD (+38.5 pp, +51.1 pp), indicating the heightened language requirements of UAV images. Secondly, the CNN-LSTM on xBD without knowledge graphs is the sole configuration in which the removal of knowledge graphs marginally enhances the CLIPScore by 4.2 percentage points, indicating that for satellite imagery with less complex damage categories, direct visual-linguistic mapping can be competitive. Appendix~\ref{app:distributions} contains the distributions of per-image scores.

\subsection{Noun-Based Object Detection Analysis}

Table~\ref{tab:noun_analysis} reports the total count of unique relevant nouns detected across all test captions to assess whether \ours{} captures more disaster-relevant objects.

\ours{} achieves full coverage of the reference object set (272 UAV objects, 640 satellite objects) compared to 55--65\% by baseline VLMs, confirming the knowledge-enriched vocabulary enables broader and more precise object references.2

\section{Discussion}
\label{sec:discussion}

\paragraph{Why knowledge graphs matter more for UAV imagery.}
The significant performance gap between with-KG and without-KG configurations on RescueNet vs. xBD shows the main differences between the two imaging methods. UAV imagery captures scenes from oblique angles at high spatial resolution. It reveals fine-grained damage details such as collapsed walls, scattered debris, and standing water that require precise vocabulary to describe. Satellite imagery, observed from a direct overhead perspective at lower resolution, shows more uniform visual patterns where more abstract descriptions suffice. RescueNet's 12-class damage taxonomy also necessitates specific terminology, which can only be provided by the knowledge-enriched vocabulary.

\paragraph{Complementarity of CLIPScore and InfoMetIC.}
Configurations with a high InfoMetIC but a moderate CLIPScore, such as the Transformer with KG on RescueNet (95.33\% vs. 73.64\%), show that using a knowledge graph helps the model include important details that aren't directly visible in the image. CLIPScore, which assesses image-text alignment, may underestimate descriptions that include domain knowledge regarding disaster processes, whereas InfoMetIC values such informational depth.

\paragraph{Practical implications.}
For operational disaster response, captions must go beyond visible elements. InfoMetIC-favored captions, which include domain-appropriate terminology and contextual reasoning, are more actionable, as demonstrated by the Transformer+KG configuration's strong performance. This suggests knowledge graphs bridge the gap between visual observation and situational understanding.

\section{Conclusion}
\label{sec:conclusion}

We introduced VLCE, a framework for generating knowledge-enriched captions for post-disaster imagery. By integrating external semantic knowledge from ConceptNet and WordNet into the caption generation pipeline, VLCE addresses the vocabulary and domain knowledge limitations of general-purpose VLMs when applied to disaster assessment. Our experiments on the xBD and RescueNet benchmarks demonstrate that knowledge graph enrichment is critical for UAV imagery captioning, where it transforms near-zero performance into strong preference over baseline VLMs. The Transformer decoder with knowledge graph integration achieves the best overall results, with 95.33\% InfoMetIC preference on RescueNet. Qualitative analysis substantiates that knowledge-enhanced captions preserve factual accuracy and contextually relevant terminology, whereas captions produced without knowledge graph assistance display hallucinations, redundancy, and semantic disparity.

\newpage
\begin{appendices}

\section{Per-Image Score Distributions}
\label{app:distributions}

This appendix provides per-image CLIPScore and InfoMetIC score distributions for all configurations. These histograms and bar charts complement the summary statistics in Table~\ref{tab:main_results}.

\subsection{CLIPScore Distributions}

\begin{figure}[H]
  \centering
  \subfloat[Score distribution]{%
    \includegraphics[width=0.45\textwidth]{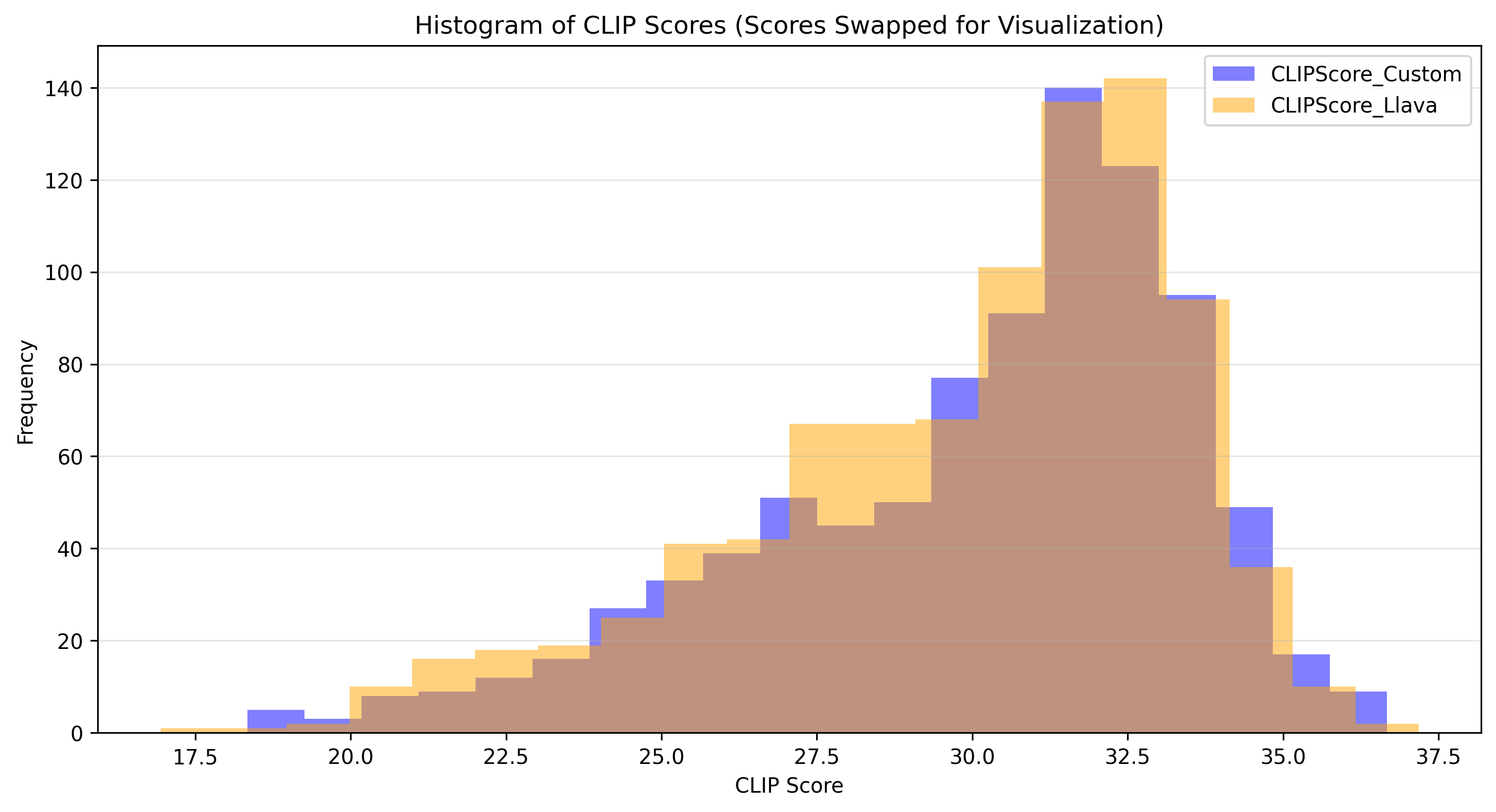}}
  \hfill
  \subfloat[Model comparison]{%
    \includegraphics[width=0.45\textwidth]{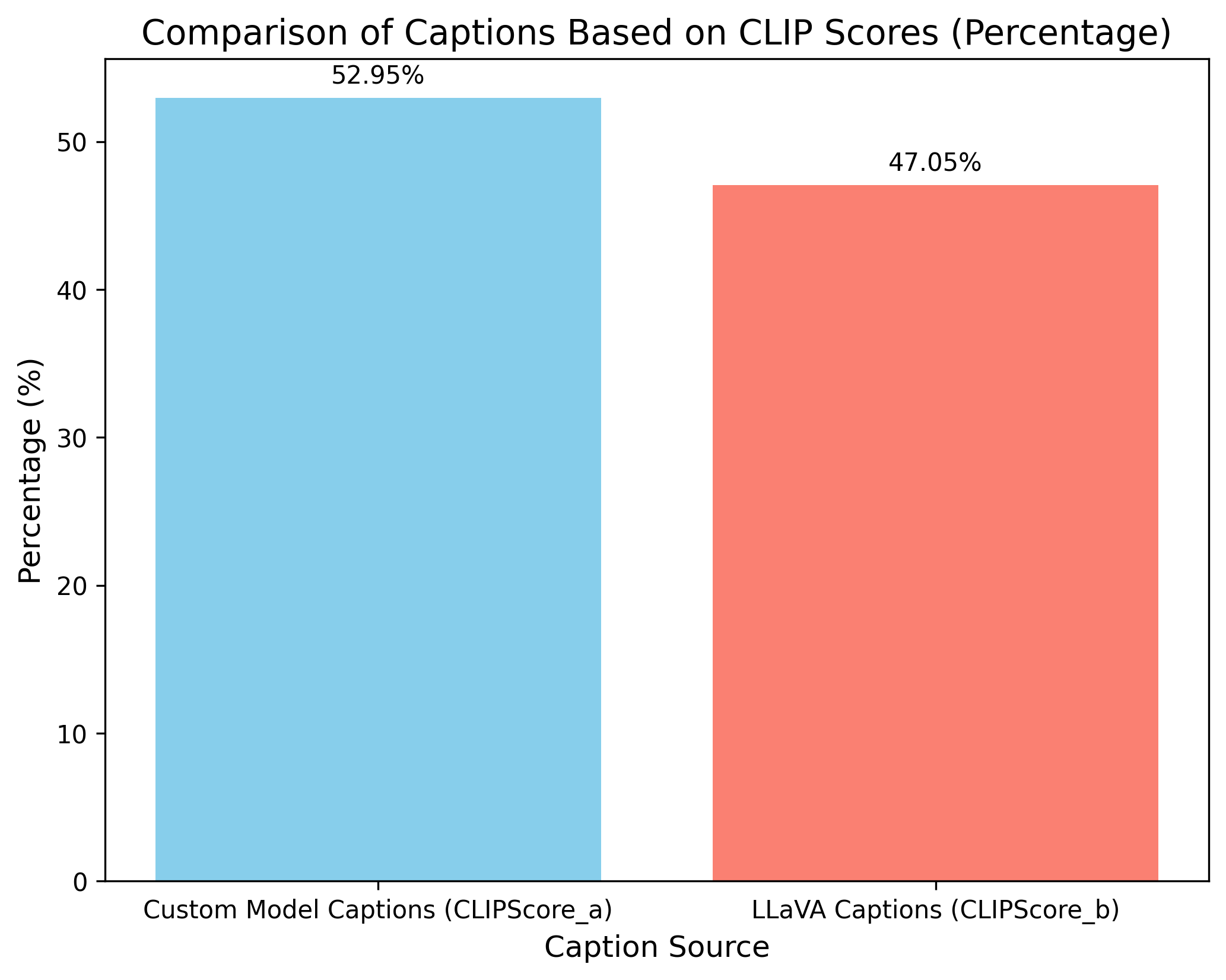}}
  \caption{CLIPScore: RescueNet + LLaVA \textbf{with} KG.}
  \label{fig:clip_rn_llava_kg}
\end{figure}

\begin{figure}[H]
  \centering
  \subfloat[Score distribution]{%
    \includegraphics[width=0.45\textwidth]{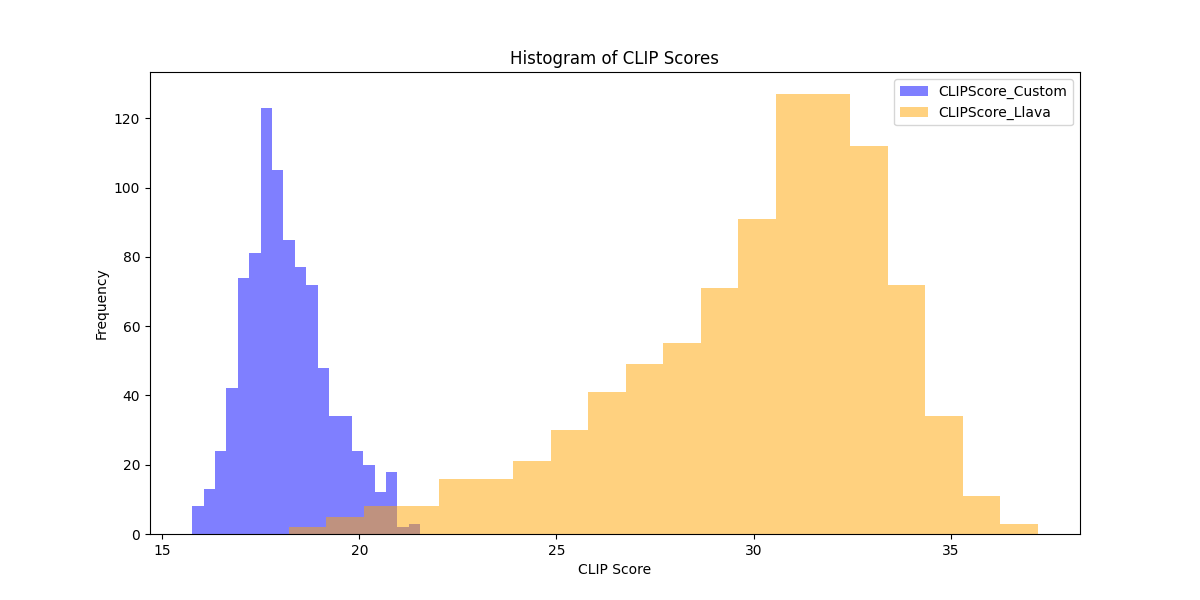}}
  \hfill
  \subfloat[Model comparison]{%
    \includegraphics[width=0.45\textwidth]{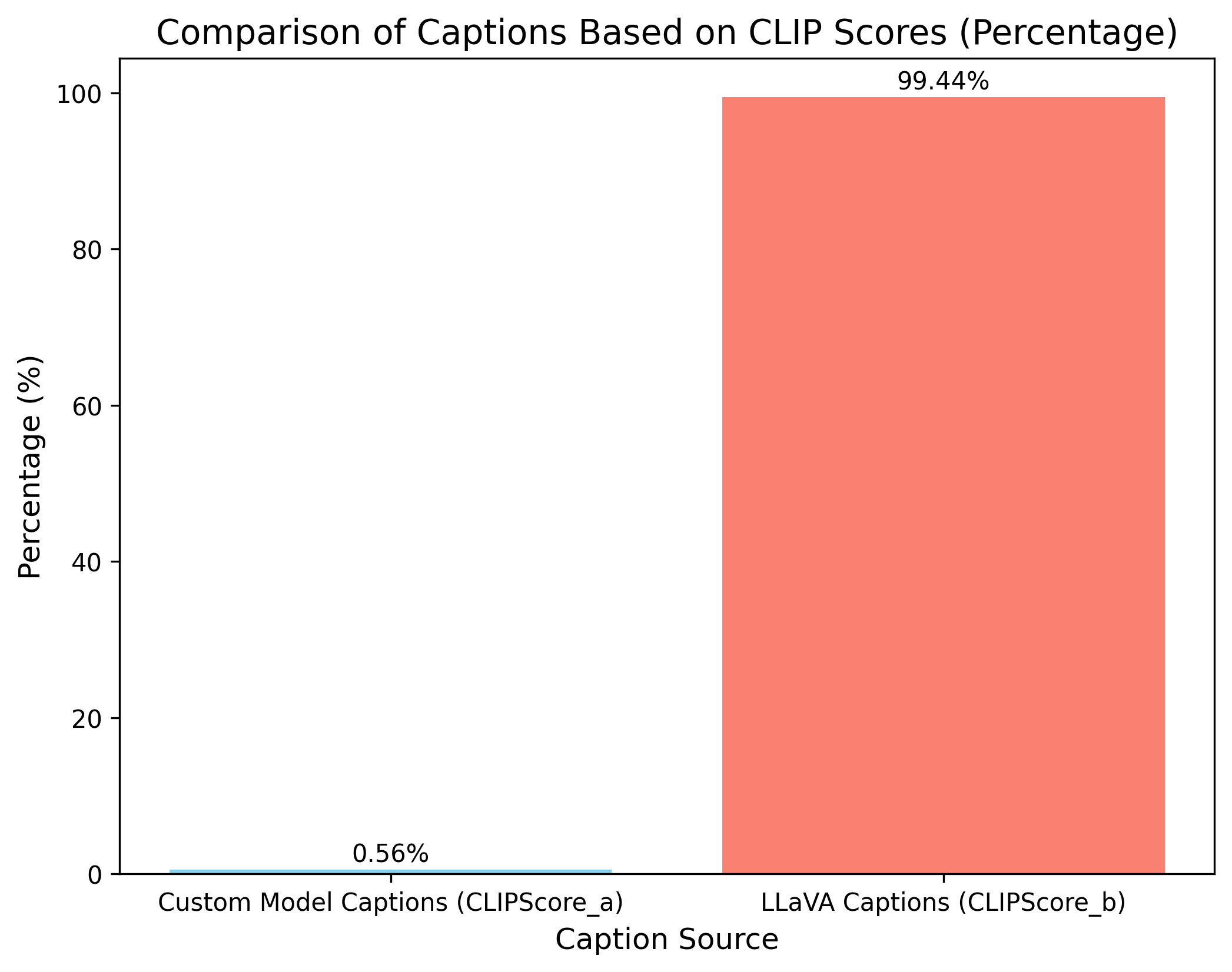}}
  \caption{CLIPScore: RescueNet + LLaVA \textbf{without} KG.}
  \label{fig:clip_rn_llava_nokg}
\end{figure}

\begin{figure}[H]
  \centering
  \subfloat[Score distribution]{%
    \includegraphics[width=0.45\textwidth]{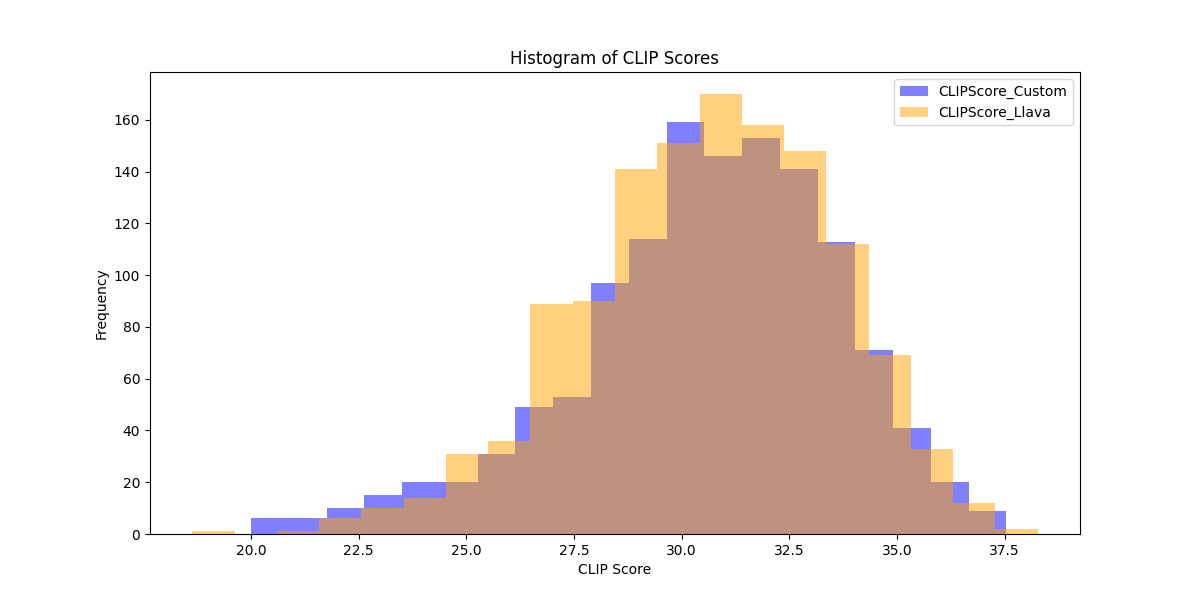}}
  \hfill
  \subfloat[Model comparison]{%
    \includegraphics[width=0.45\textwidth]{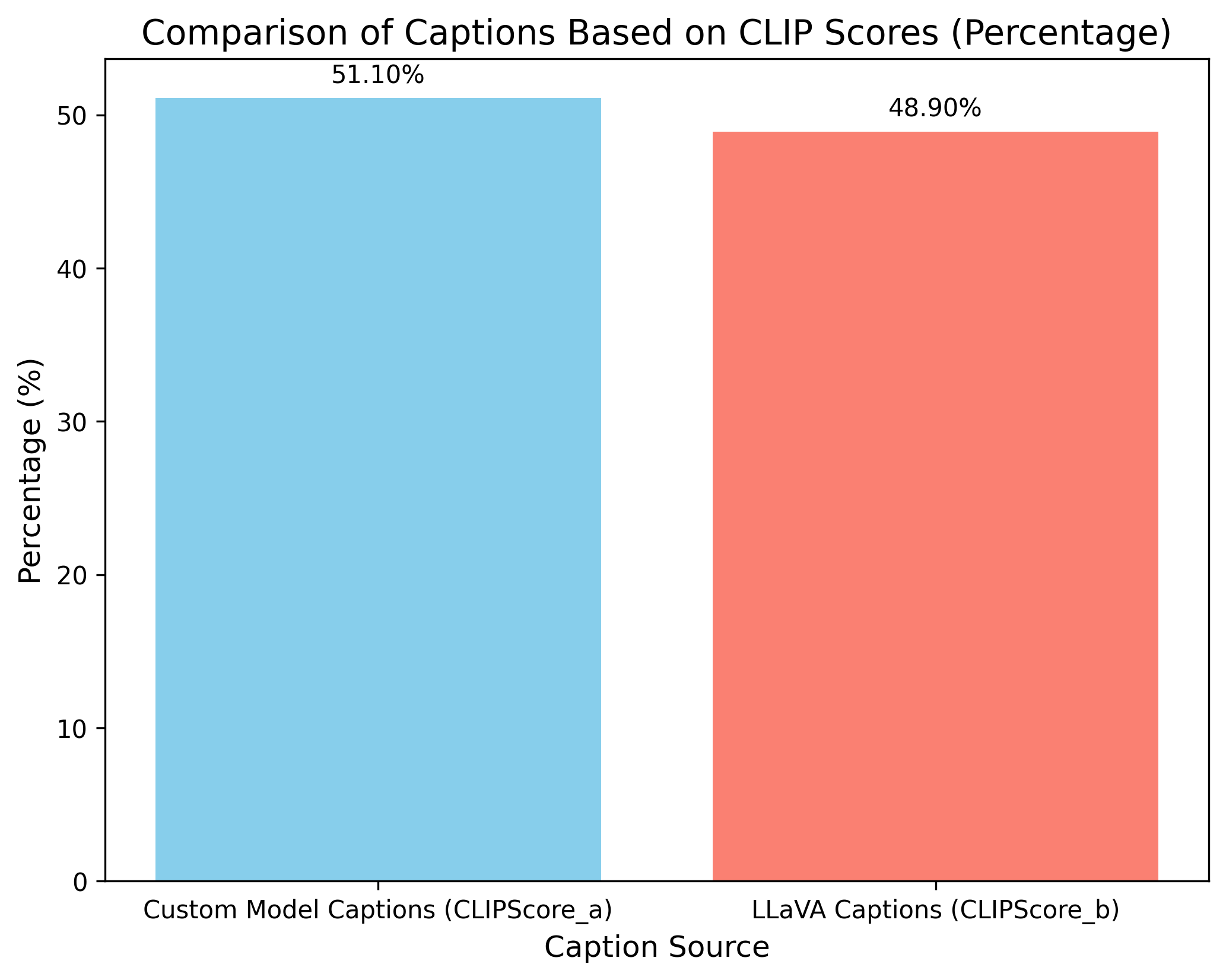}}
  \caption{CLIPScore: xBD + LLaVA \textbf{with} KG.}
  \label{fig:clip_xbd_llava_kg}
\end{figure}

\begin{figure}[H]
  \centering
  \subfloat[Score distribution]{%
    \includegraphics[width=0.45\textwidth]{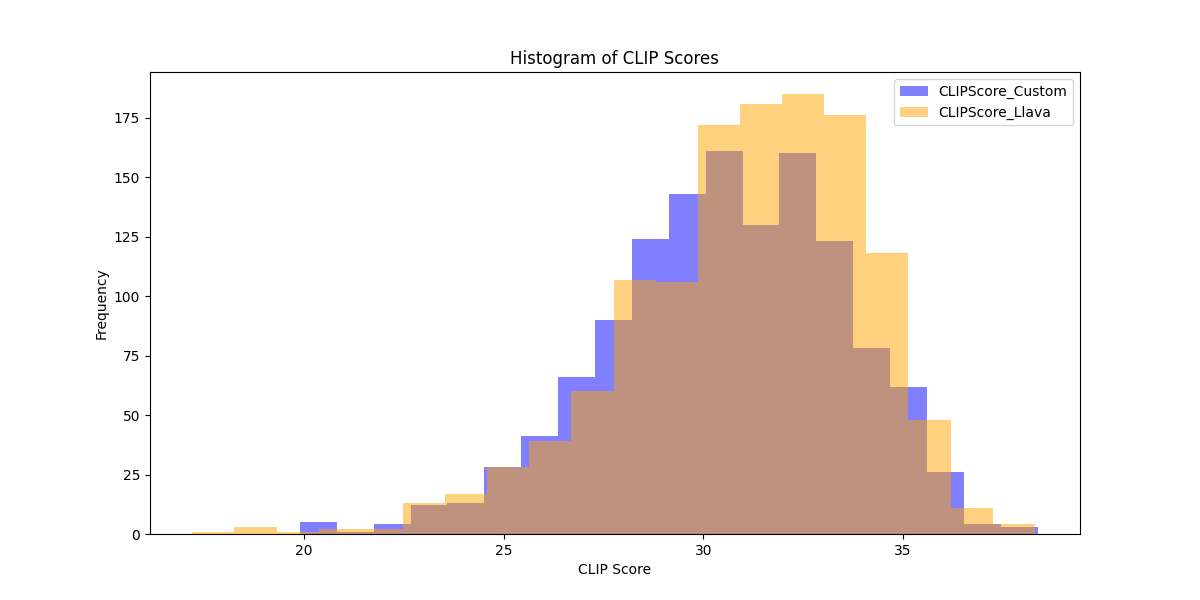}}
  \hfill
  \subfloat[Model comparison]{%
    \includegraphics[width=0.45\textwidth]{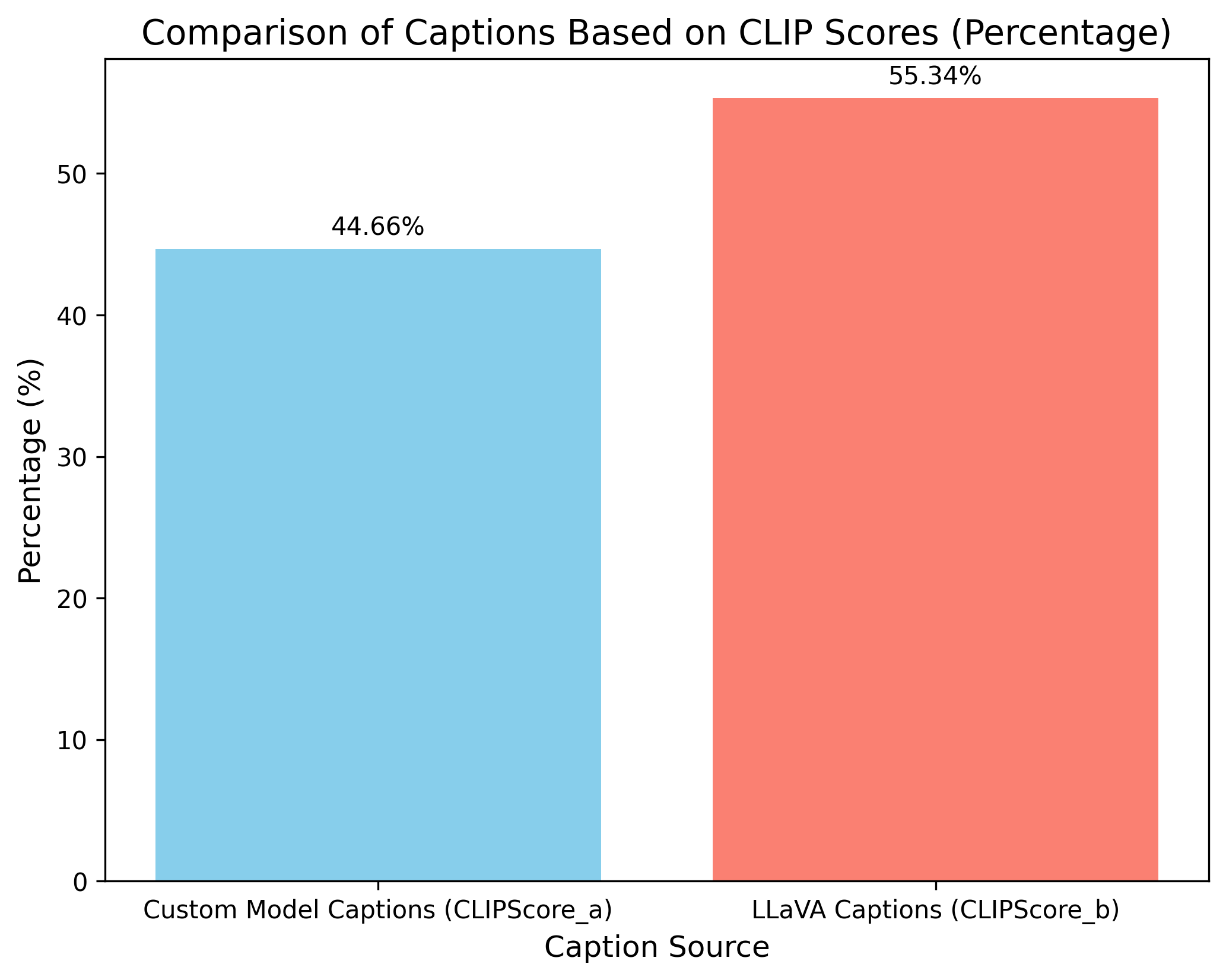}}
  \caption{CLIPScore: xBD + LLaVA \textbf{without} KG.}
  \label{fig:clip_xbd_llava_nokg}
\end{figure}

\begin{figure}[H]
  \centering
  \subfloat[Score distribution]{%
    \includegraphics[width=0.45\textwidth]{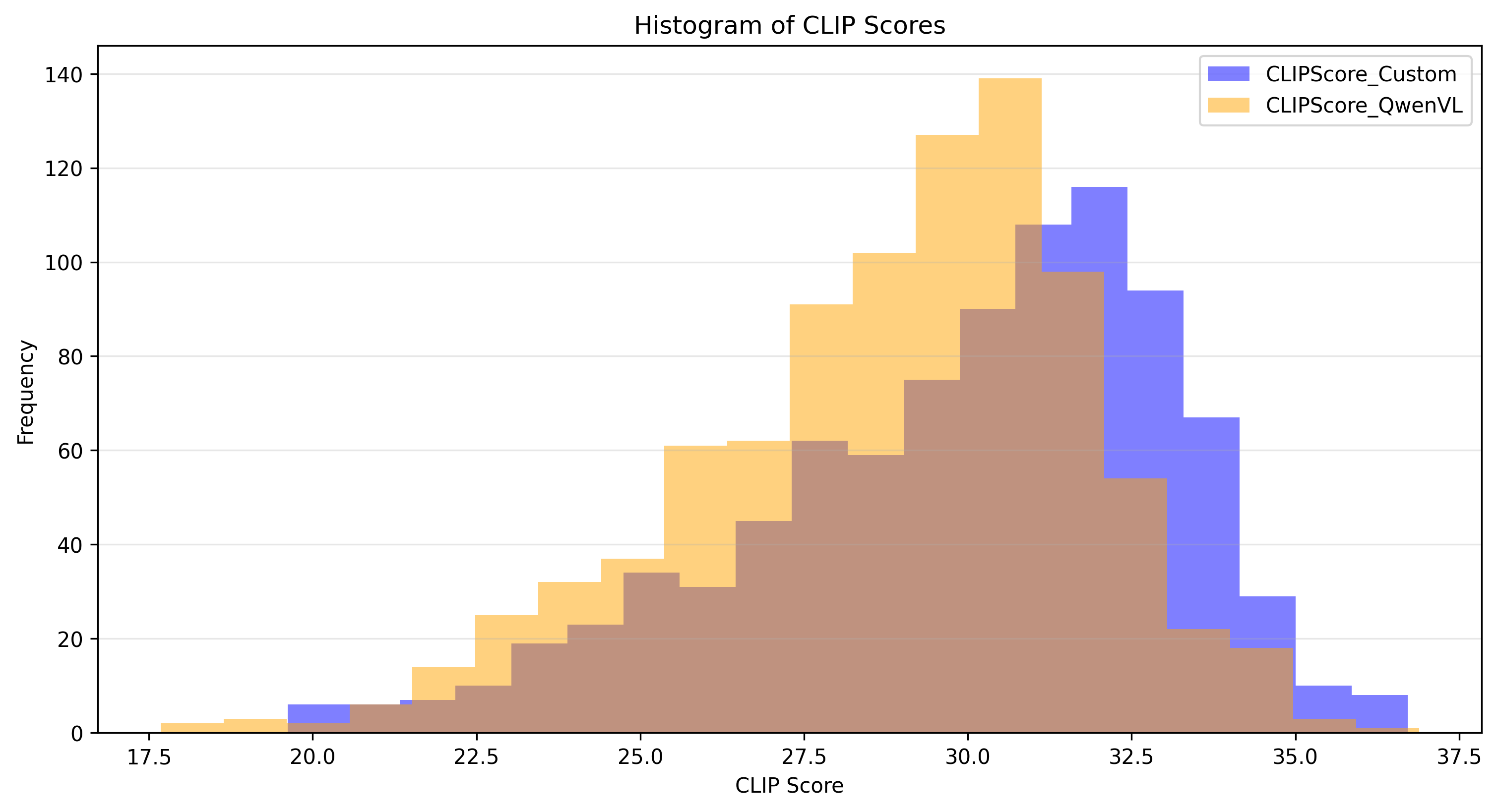}}
  \hfill
  \subfloat[Model comparison]{%
    \includegraphics[width=0.45\textwidth]{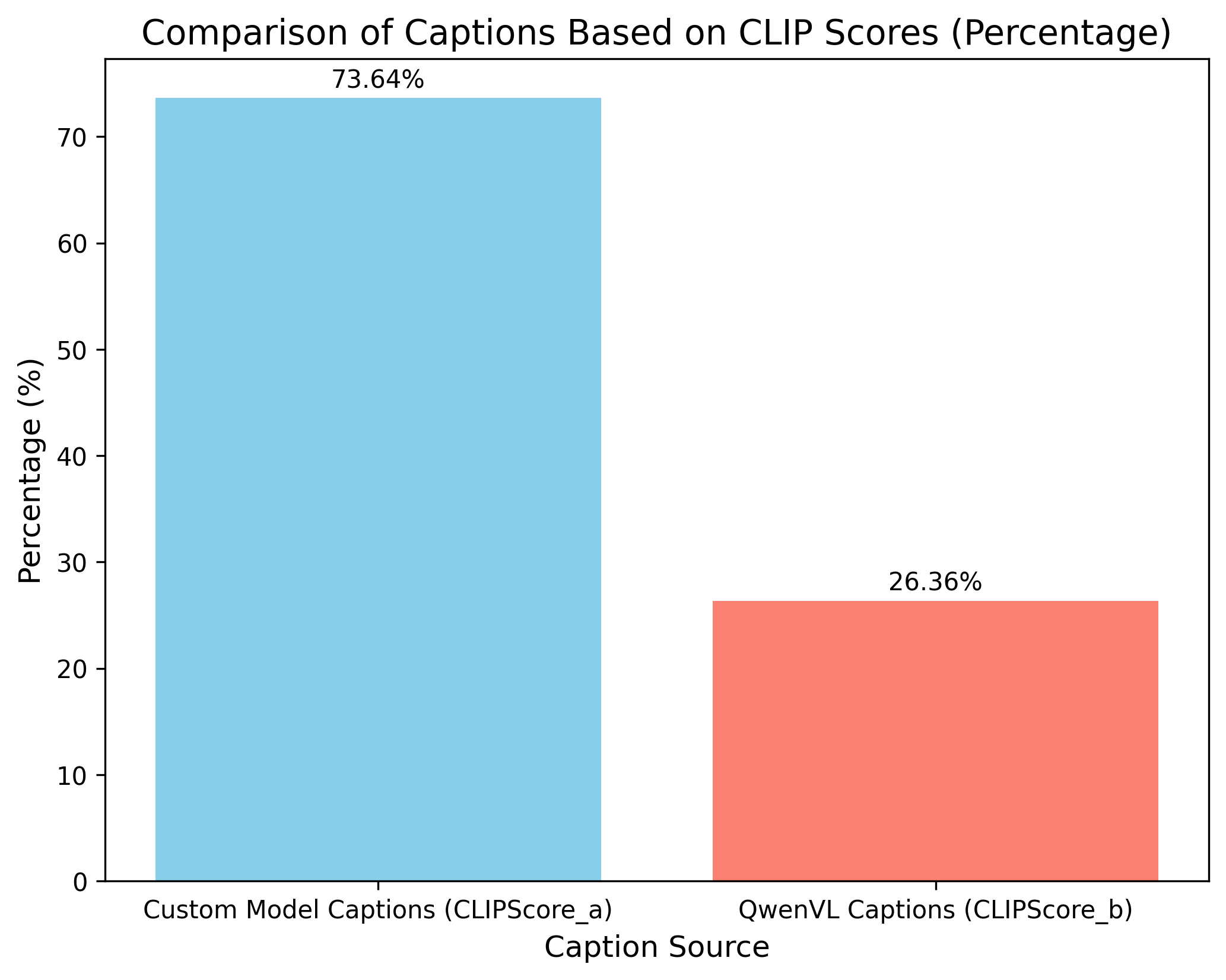}}
  \caption{CLIPScore: RescueNet + QwenVL \textbf{with} KG.}
  \label{fig:clip_rn_qwen_kg}
\end{figure}

\begin{figure}[H]
  \centering
  \subfloat[Score distribution]{%
    \includegraphics[width=0.45\textwidth]{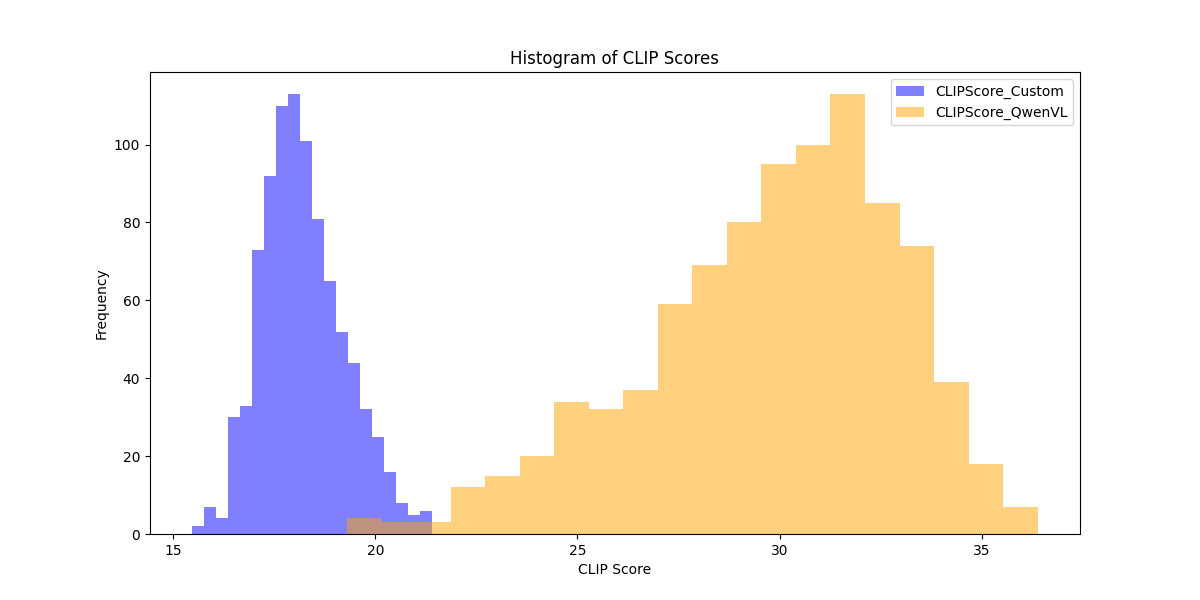}}
  \hfill
  \subfloat[Model comparison]{%
    \includegraphics[width=0.45\textwidth]{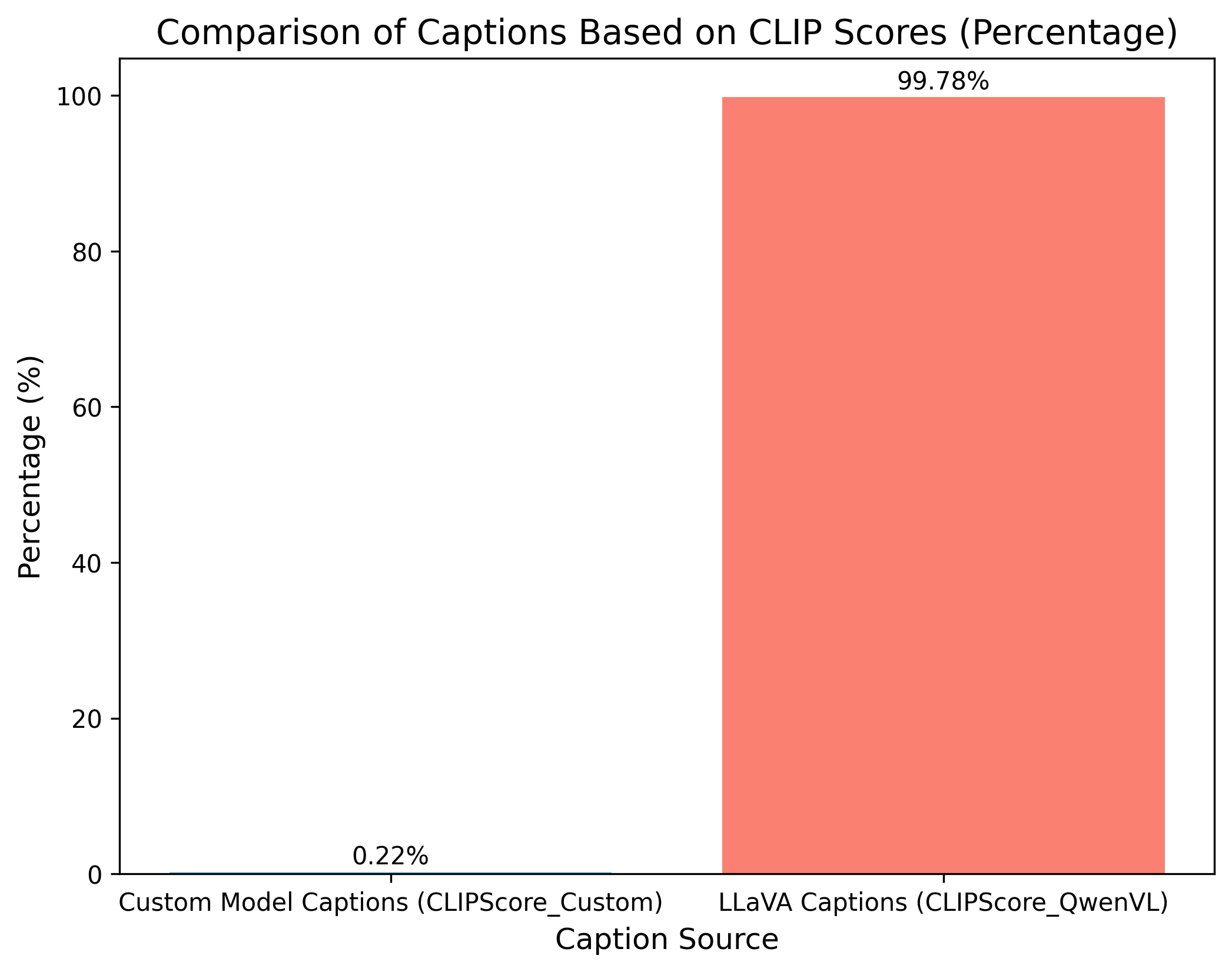}}
  \caption{CLIPScore: RescueNet + QwenVL \textbf{without} KG.}
  \label{fig:clip_rn_qwen_nokg}
\end{figure}

\begin{figure}[H]
  \centering
  \subfloat[With KG]{%
    \includegraphics[width=0.45\textwidth]{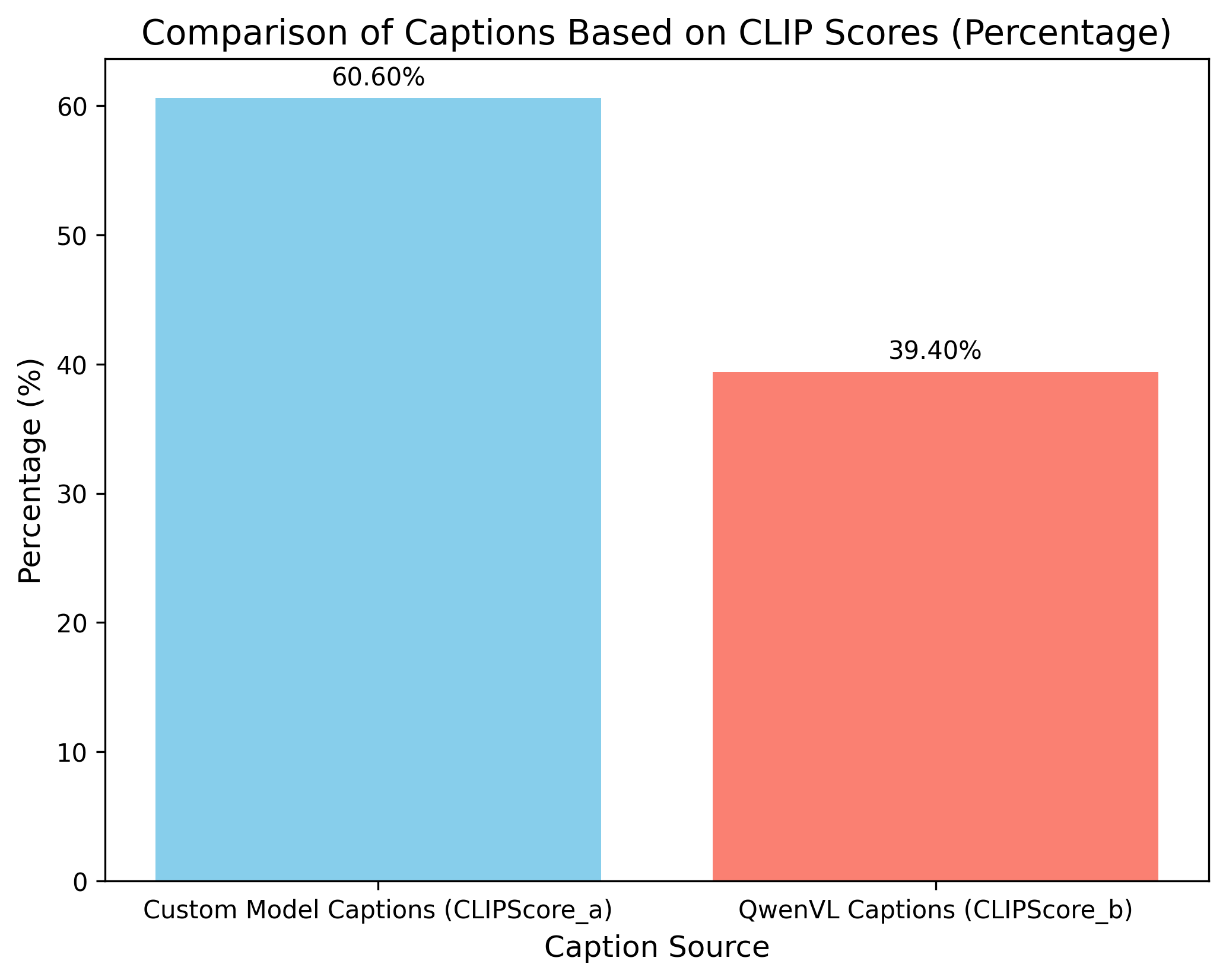}}
  \hfill
  \subfloat[Without KG]{%
    \includegraphics[width=0.45\textwidth]{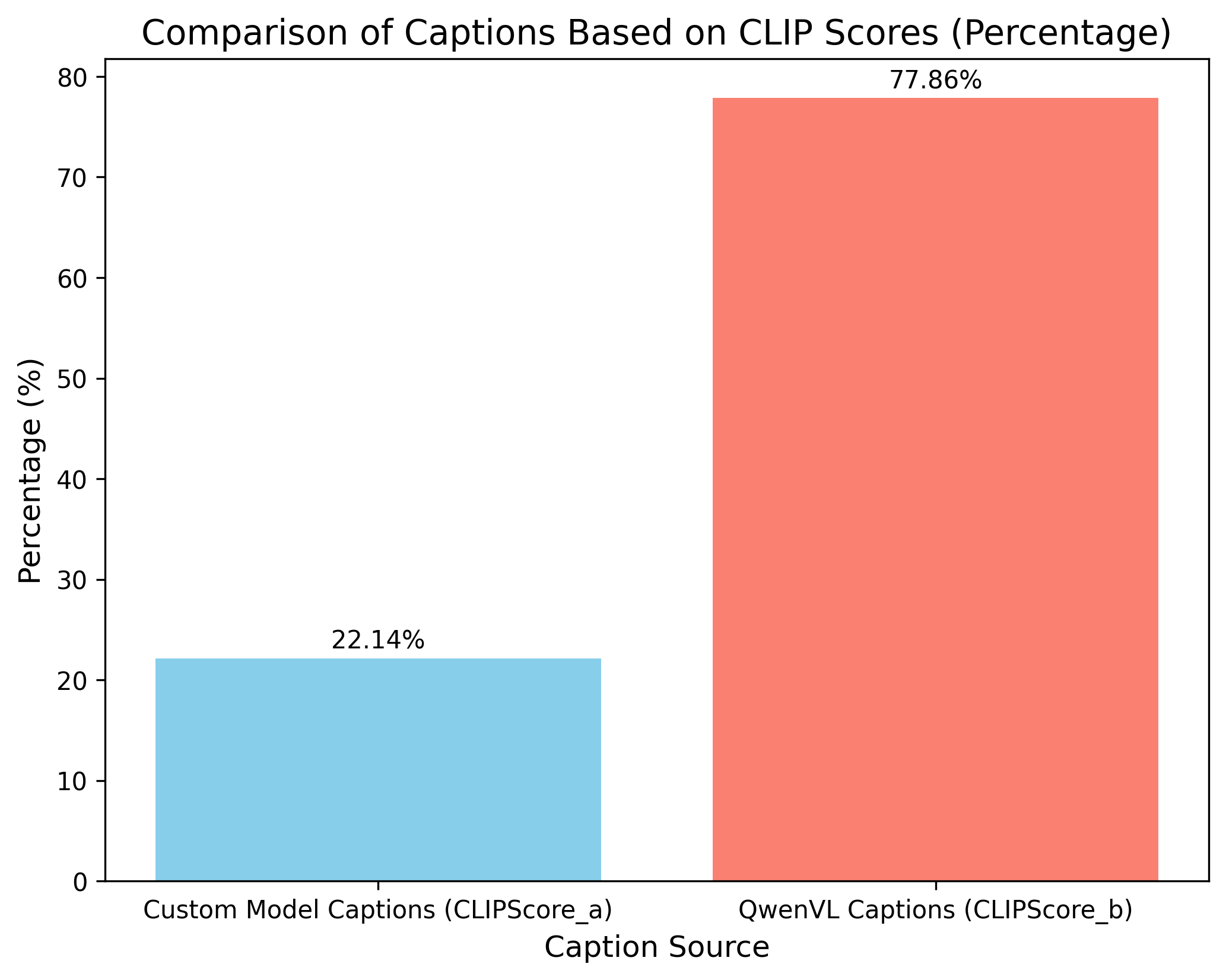}}
  \caption{CLIPScore: xBD + QwenVL, with vs.\ without KG.}
  \label{fig:clip_xbd_qwen_both}
\end{figure}

\subsection{InfoMetIC Distributions}

\begin{figure}[H]
  \centering
  \subfloat[Model comparison]{%
    \includegraphics[width=0.45\textwidth]{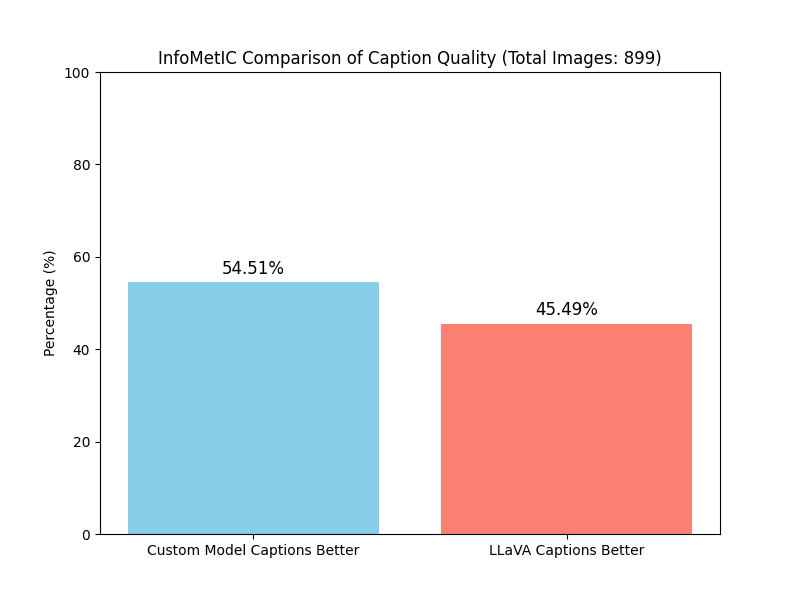}}
  \hfill
  \subfloat[Score distribution]{%
    \includegraphics[width=0.45\textwidth]{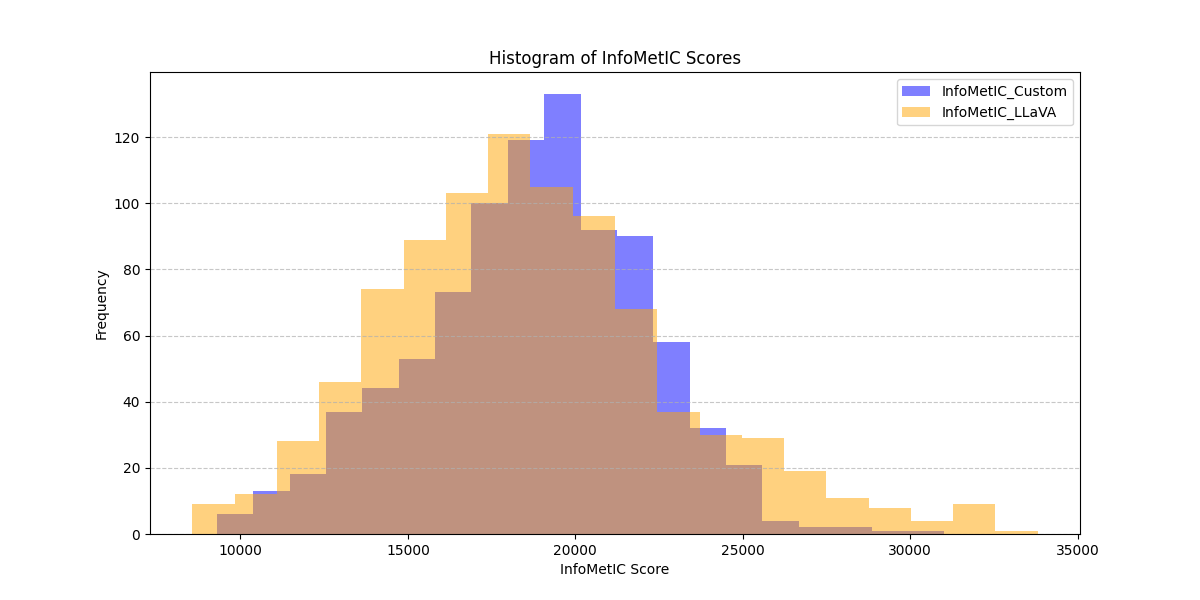}}
  \caption{InfoMetIC: RescueNet + LLaVA \textbf{with} KG.}
  \label{fig:info_rn_llava_kg}
\end{figure}

\begin{figure}[H]
  \centering
  \subfloat[Model comparison]{%
    \includegraphics[width=0.45\textwidth]{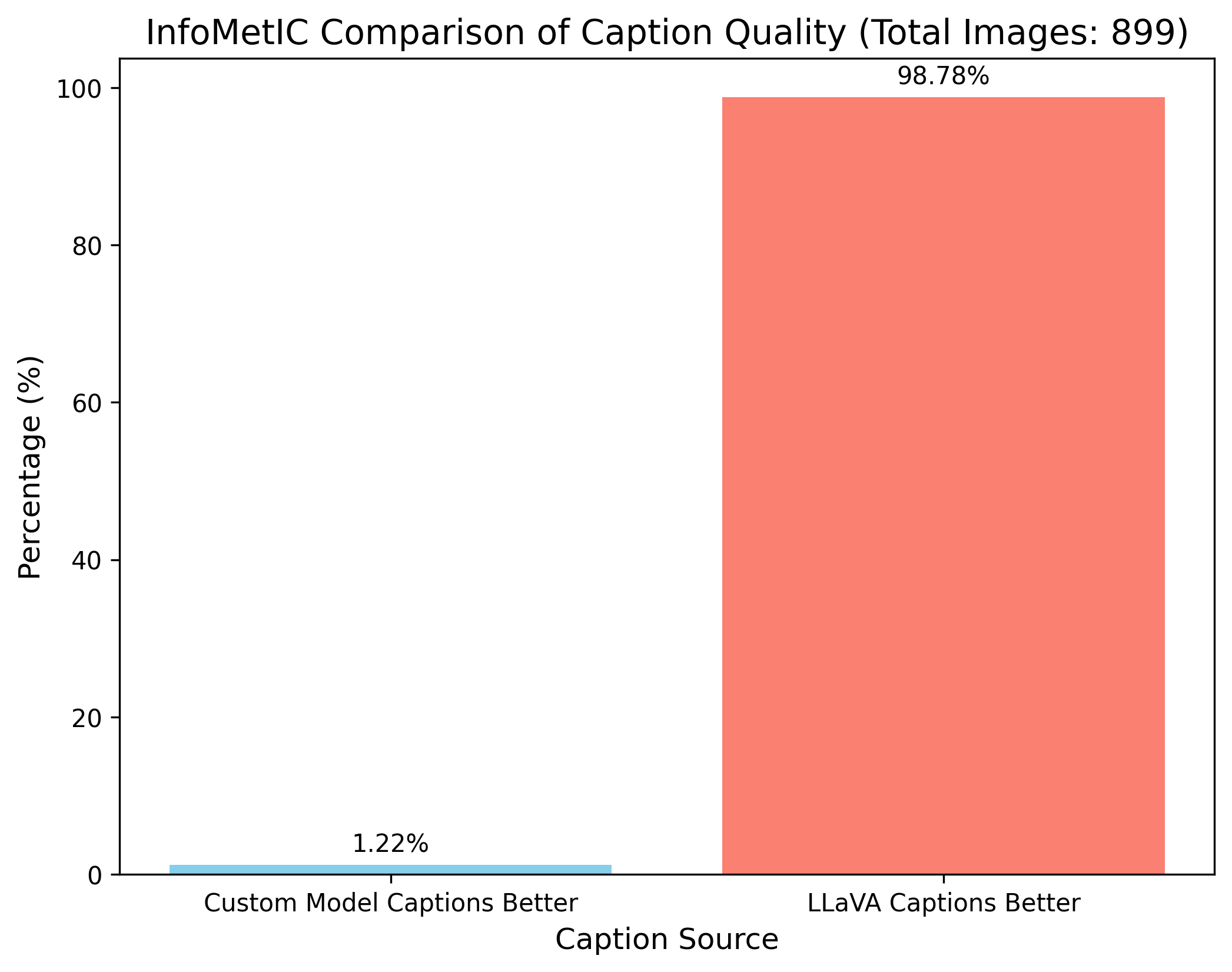}}
  \hfill
  \subfloat[Score distribution]{%
    \includegraphics[width=0.45\textwidth]{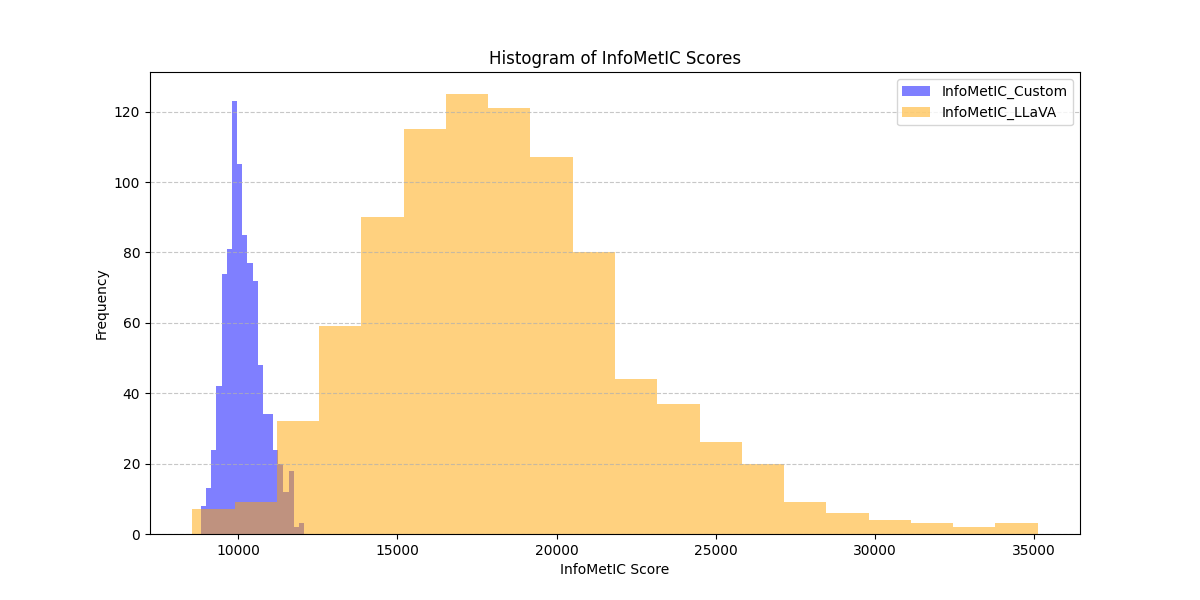}}
  \caption{InfoMetIC: RescueNet + LLaVA \textbf{without} KG.}
  \label{fig:info_rn_llava_nokg}
\end{figure}

\begin{figure}[H]
  \centering
  \subfloat[Model comparison]{%
    \includegraphics[width=0.45\textwidth]{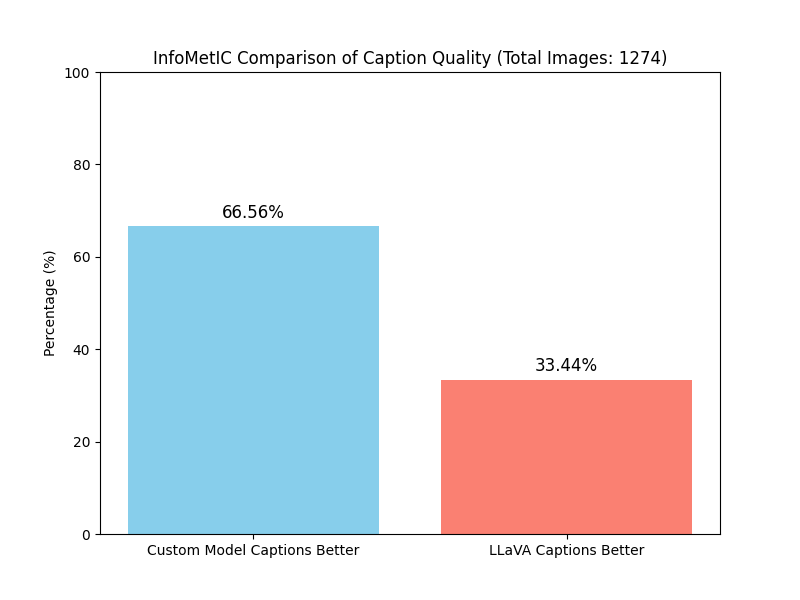}}
  \hfill
  \subfloat[Score distribution]{%
    \includegraphics[width=0.45\textwidth]{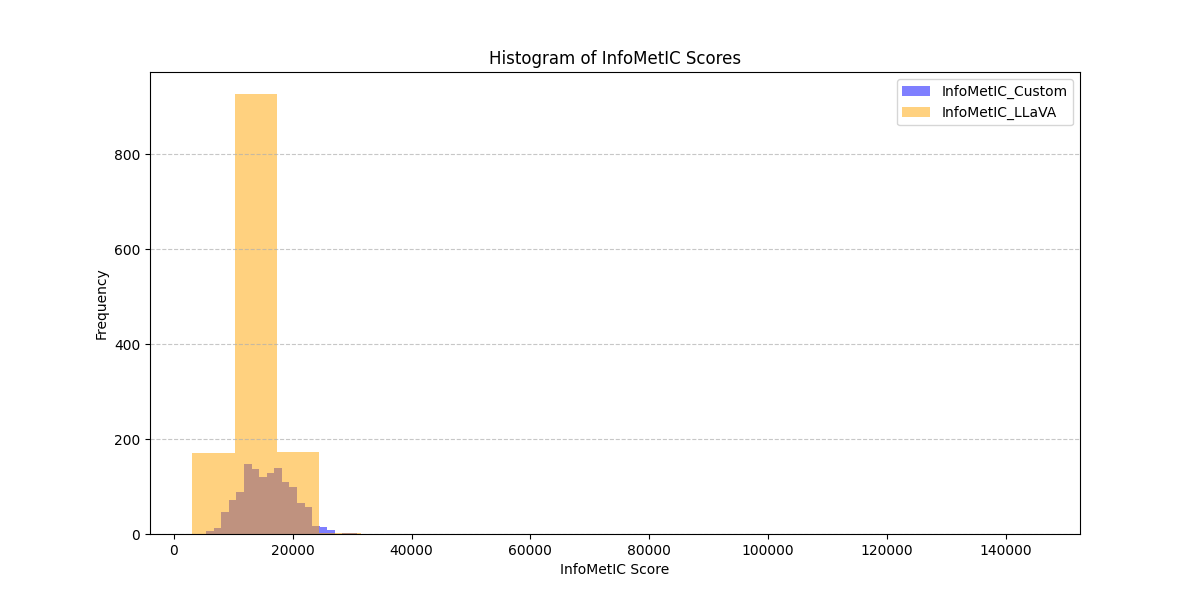}}
  \caption{InfoMetIC: xBD + LLaVA \textbf{with} KG.}
  \label{fig:info_xbd_llava_kg}
\end{figure}

\begin{figure}[H]
  \centering
  \subfloat[Model comparison]{%
    \includegraphics[width=0.45\textwidth]{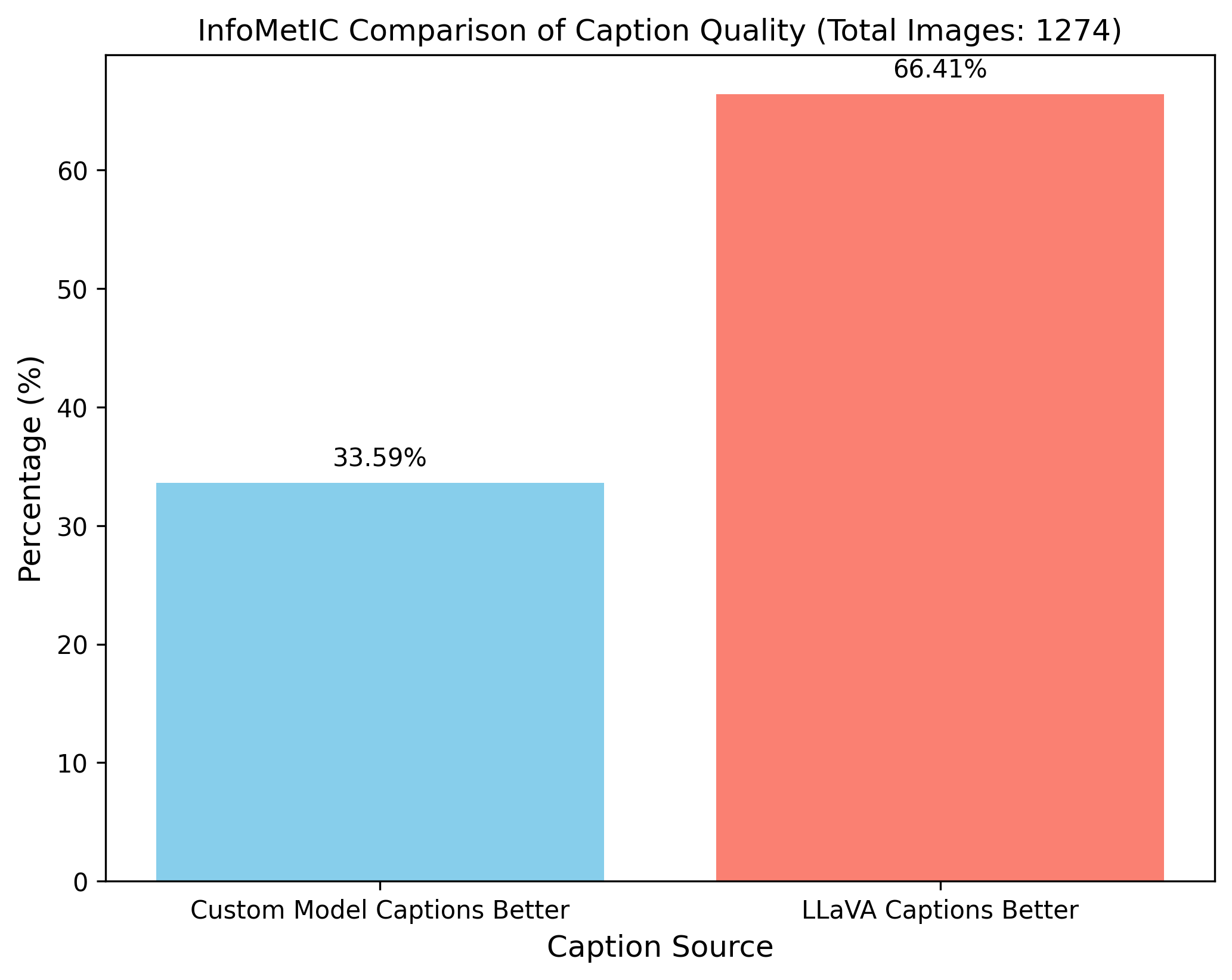}}
  \hfill
  \subfloat[Score distribution]{%
    \includegraphics[width=0.45\textwidth]{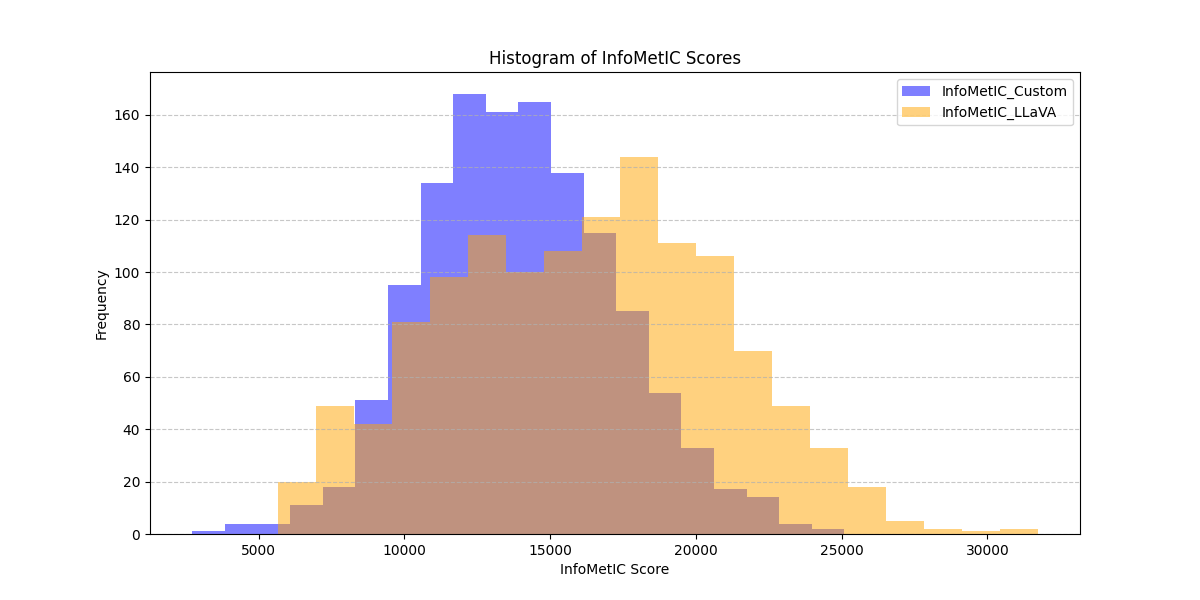}}
  \caption{InfoMetIC: xBD + LLaVA \textbf{without} KG.}
  \label{fig:info_xbd_llava_nokg}
\end{figure}

\begin{figure}[H]
  \centering
  \subfloat[Model comparison]{%
    \includegraphics[width=0.45\textwidth]{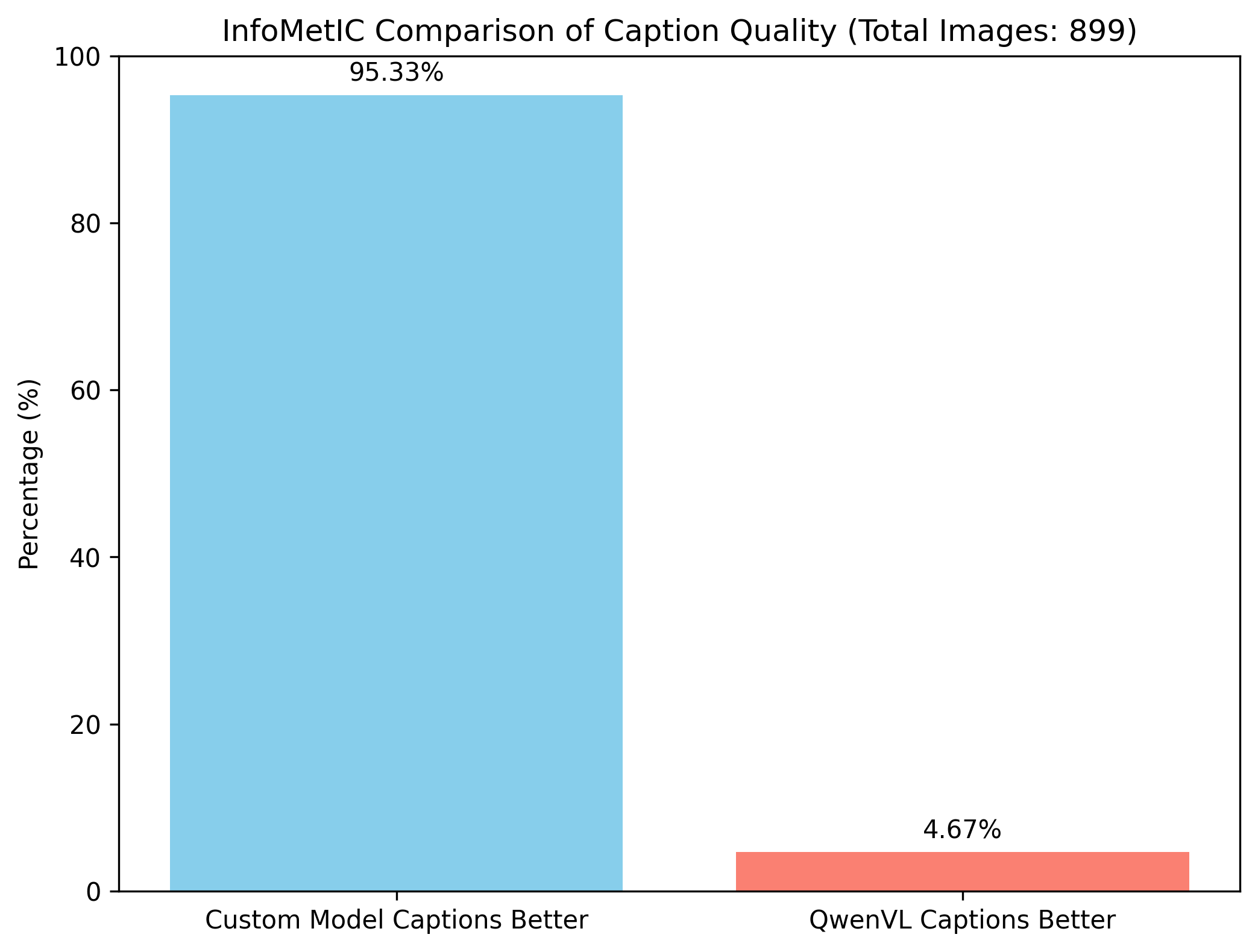}}
  \hfill
  \subfloat[Score distribution]{%
    \includegraphics[width=0.45\textwidth]{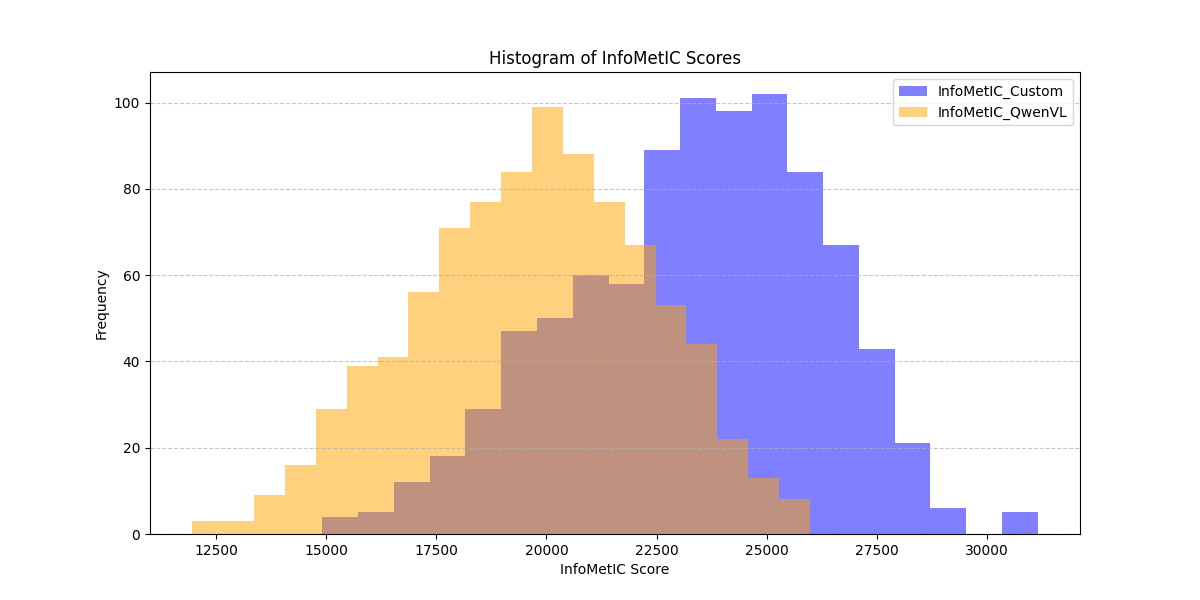}}
  \caption{InfoMetIC: RescueNet + QwenVL \textbf{with} KG.}
  \label{fig:info_rn_qwen_kg}
\end{figure}

\begin{figure}[H]
  \centering
  \subfloat[Model comparison]{%
    \includegraphics[width=0.45\textwidth]{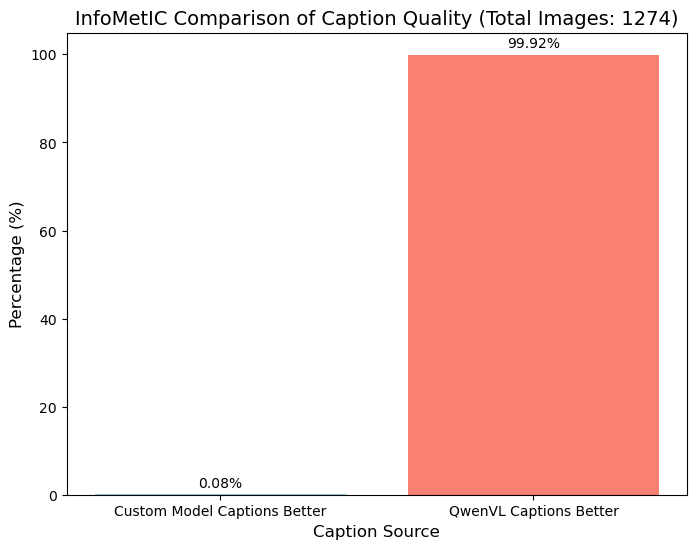}}
  \hfill
  \subfloat[Score distribution]{%
    \includegraphics[width=0.45\textwidth]{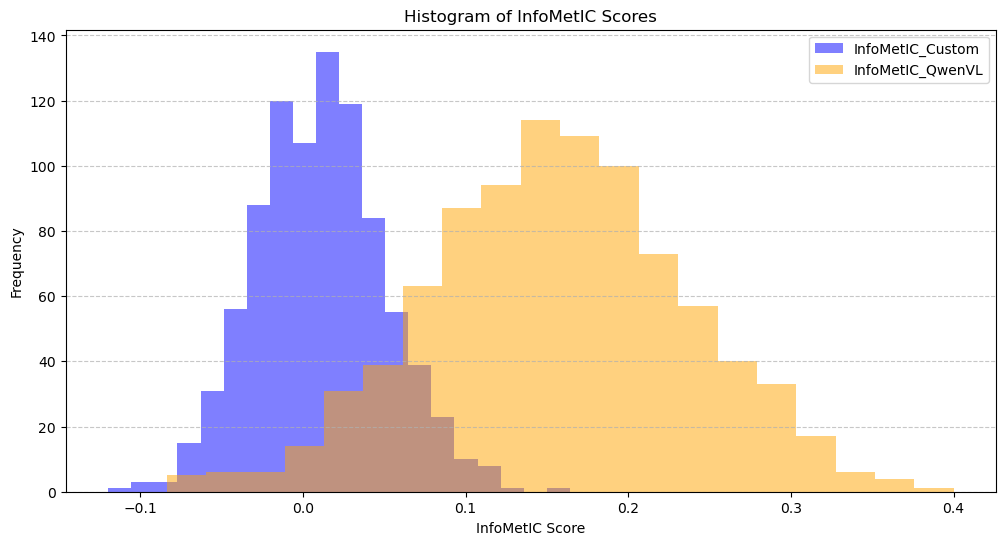}}
  \caption{InfoMetIC: RescueNet + QwenVL \textbf{without} KG.}
  \label{fig:info_rn_qwen_nokg}
\end{figure}

\begin{figure}[H]
  \centering
  \subfloat[With KG]{%
    \includegraphics[width=0.45\textwidth]{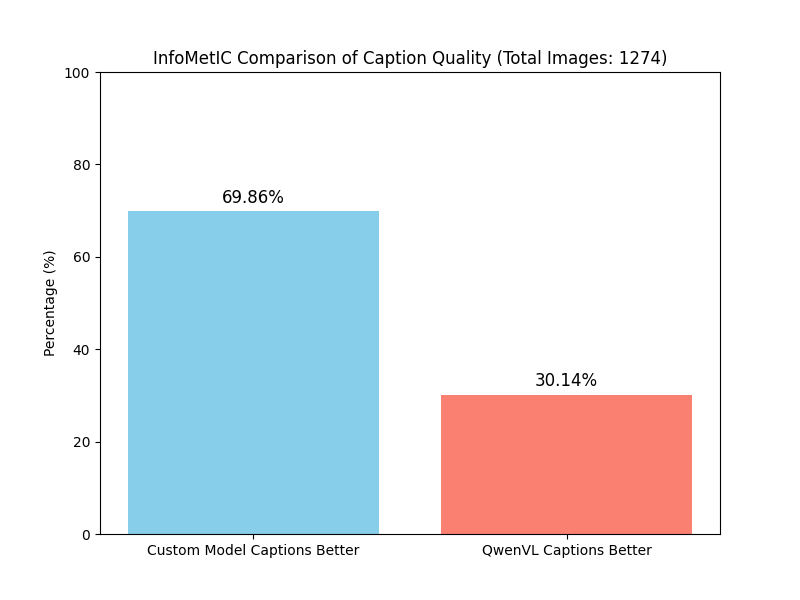}}
  \hfill
  \subfloat[Without KG]{%
    \includegraphics[width=0.45\textwidth]{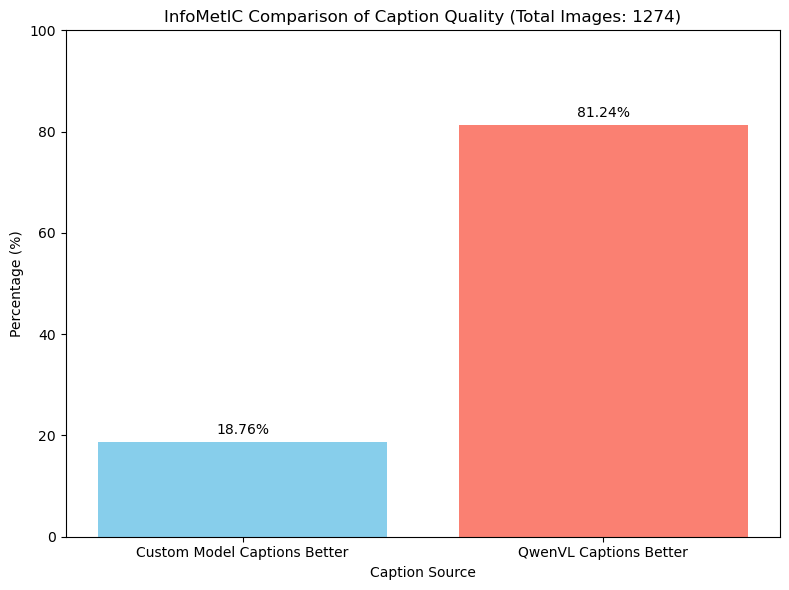}}
  \caption{InfoMetIC: xBD + QwenVL, with vs.\ without KG.}
  \label{fig:info_xbd_qwen_both}
\end{figure}

\section{Qualitative Case Studies}
\label{app:qualitative}

We examine representative examples across dataset-model combinations to illustrate how KG integration affects caption quality.

\subsection{RescueNet + LLaVA}

\begin{figure}[h!]
    \centering
    \includegraphics[width=0.60\textwidth]{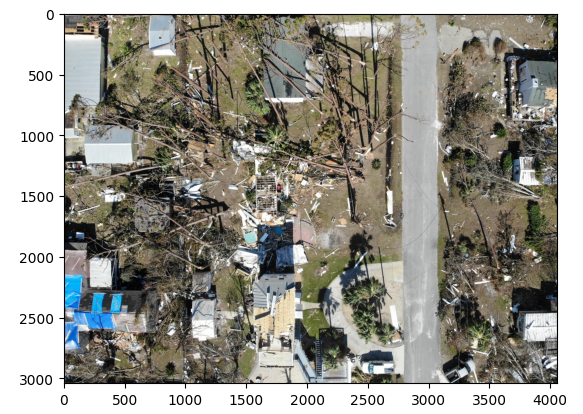}
    \caption{RescueNet with KG-enhanced LLaVA output.}
    \label{fig:example1}
\end{figure}

\begin{tcolorbox}[colback=gray!5, colframe=black!50, coltitle=black, fonttitle=\bfseries, title=KG-Enhanced Caption {[\ref{fig:example1}]}]
\textbf{Caption:} the image depicts the aftermath of hurricane michael which caused significant damage to the infrastructure and environment in the affected area \textcolor{green}{\textbf{the most noticeable feature is the flooded area which has submerged the debris from damaged structures can be seen scattered throughout the scene}} with some debris piled up in the water the presence of multiple vehicles including cars and trucks indicates that the area is being cleared and assessed for recovery efforts the overall scene conveys sense of destruction and recovery efforts needed to restore the affected area to its previous state.
\end{tcolorbox}

The KG-enhanced caption coherently identifies flooding as the primary damage type, notes submerged debris, and infers recovery operations from visible vehicles.

\begin{figure}[h!]
    \centering
    \includegraphics[width=0.55\textwidth]{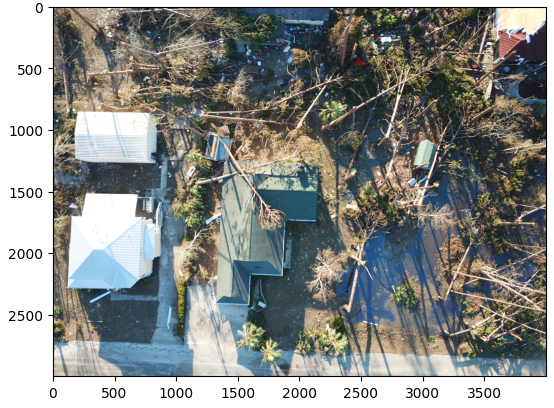}
    \caption{RescueNet without KG, LLaVA output.}
    \label{fig:example2}
\end{figure}

\begin{tcolorbox}[colback=gray!5, colframe=black!50, coltitle=black, fonttitle=\bfseries, title=Baseline Caption (No KG) {[\ref{fig:example2}]}]
\textbf{Caption:} \textcolor{purple}{\textbf{the the}} image shows an aerial view of a residential severely impacted by a natural disaster fallen trees lie across rooftops, \textcolor{purple}{\textbf{stone}} and streets directly visible structural damage can be seen while the surrounding area is littered with broken branches and debris the stark contrast between intact rooftops and \textcolor{purple}{\textbf{widespreadvegetation}} damage impacted \textcolor{purple}{\textbf{by a natural disaster fallen trees lie rooftops, stone}}
\end{tcolorbox}

Without KG: word duplication (``the the''), nonsensical terms (``stone''), concatenated words (``widespreadvegetation''), and verbatim phrase repetition.

\subsection{xBD + LLaVA}

\begin{figure}[h!]
    \centering
    \includegraphics[width=0.55\textwidth]{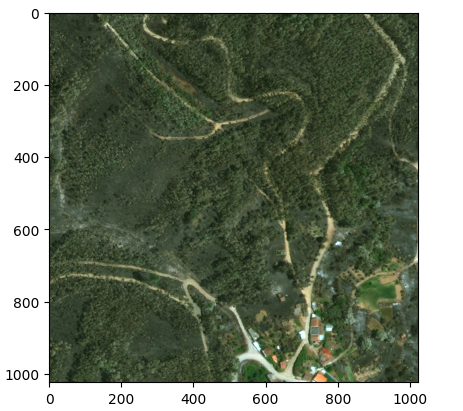}
    \caption{xBD with KG-enhanced LLaVA output.}
    \label{fig:example3}
\end{figure}

\begin{tcolorbox}[colback=gray!5, colframe=black!50, coltitle=black, fonttitle=\bfseries, title=KG-Enhanced Caption {[\ref{fig:example3}]}]
\textbf{Caption:} the image shows satellite view of forest with clear path through the trees the path is surrounded by trees and there are several trees in the area the pathway appears to be dirt road and it is surrounded by trees the \textcolor{green}{\textbf{image also shows few buildings in the area which seem to have escaped the aftermath of the disaster}} the image captures the aftermath of disaster with undamaged areas or buildings still standing amidst the destruction
\end{tcolorbox}

\begin{figure}[h!]
    \centering
    \includegraphics[width=0.55\textwidth]{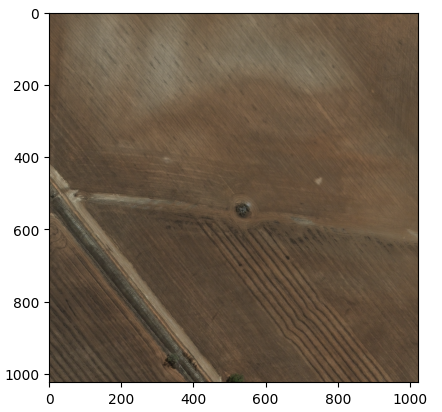}
    \caption{xBD without KG, LLaVA output.}
    \label{fig:example4}
\end{figure}

\begin{tcolorbox}[colback=gray!5, colframe=black!50, coltitle=black, fonttitle=\bfseries, title=Baseline Caption (No KG) {[\ref{fig:example4}]}]
\textbf{Caption:} the image shows satellite view of field with large crack in the middle likely caused by an earthquake \textcolor{purple}{\textbf{the crack extends from the left side of the image to the right dividing the field into two distinct sections additionally there are two people in the field one near the center}} and the other towards the right side the presence of these holes suggests that the area has been affected by the disaster and\textcolor{purple}{\textbf{the animals deaths may have been temporarily closed due to the damage}}
\end{tcolorbox}

The baseline fabricates elements (people in the field, animal deaths) and produces semantically incoherent phrases---a dangerous failure mode for disaster assessment.

\subsection{RescueNet + QwenVL}

\begin{figure}[h!]
    \centering
    \includegraphics[width=0.55\textwidth]{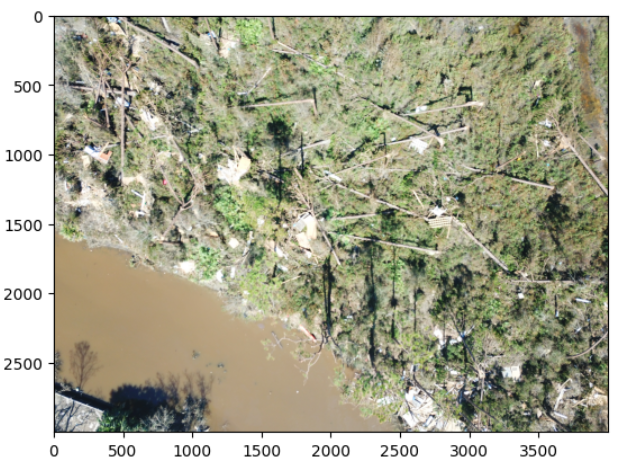}
    \caption{RescueNet with KG-enhanced QwenVL output.}
    \label{fig:example5}
\end{figure}

\begin{tcolorbox}[colback=gray!5, colframe=black!50, coltitle=black, fonttitle=\bfseries, title=KG-Enhanced Caption {[\ref{fig:example5}]}]
\textbf{Caption:} the satellite image depicts an area that has been significantly impacted by hurricane the \textcolor{green}{\textbf{image shows dense concentration of trees that appear to be damaged or fallen indicating the powerful force of the storm the trees are scattered}} throughout the area with some lying on the ground and others leaning precariously \textcolor{green}{\textbf{the extent of the destruction is widespread affecting both the natural landscape}} and built environments the fallen trees and debris suggest that \textcolor{green}{\textbf{the hurricane had strong impact on the vegetation potentially causing widespread damage to the local ecosystem}} the presence of debris and fallen trees can obstruct roads and paths making it difficult for emergency services and residents 
\end{tcolorbox}

\begin{figure}[h!]
    \centering
    \includegraphics[width=0.55\textwidth]{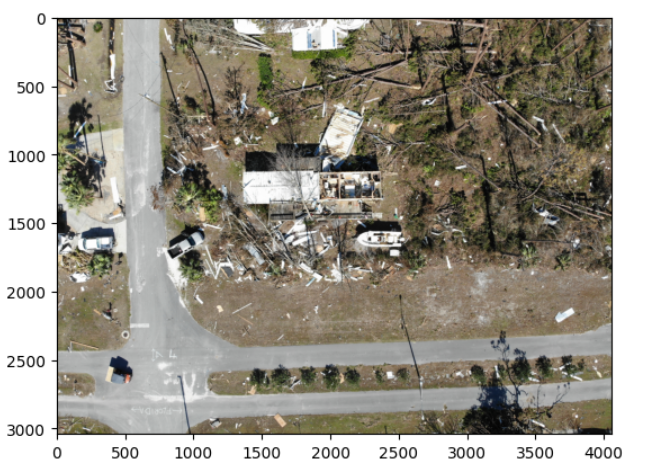}
    \caption{RescueNet without KG, QwenVL output.}
    \label{fig:example6}
\end{figure}

\begin{tcolorbox}[colback=gray!5, colframe=black!50, coltitle=black, fonttitle=\bfseries, title=Baseline Caption (No KG) {[\ref{fig:example6}]}]
\textbf{Caption:} the satellite photo depicts an area that has been significantly impacted by \textcolor{purple}{\textbf{a food area and causing harm to five people}} of the destruction is evident from the scattered debris and damaged structures here is detailed description of the visible damage and its implications damaged structures there are two large piles of debris scattered across the area these piles consist of various materials such as trees, wood metal and possibly concrete indicating that multiple buildings have been severely damaged or destroyed the debris 
\end{tcolorbox}

The baseline generates semantic inaccuracies (``food area and causing harm to five people''), fabricated casualty figures, and fragmented sentences.

\subsection{xBD + QwenVL}

\begin{figure}[h!]
  \centering
  \includegraphics[width=0.55\textwidth]{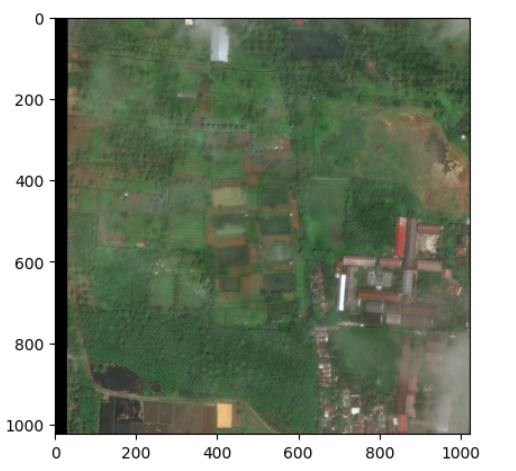}
  \caption{xBD with KG-enhanced QwenVL output.}
  \label{fig:example7a}
\end{figure}

\begin{tcolorbox}[colback=gray!5, colframe=black!50, coltitle=black, fonttitle=\bfseries, title=KG-Enhanced Caption {[\ref{fig:example7a}]}]
\textbf{Caption:} The image shows \textcolor{green}{\textbf{satellite view of town with mix of undamaged and damaged buildings there are undamaged buildings and damaged buildings visible in the scene}} the presence of both undamaged and damaged structures highlights the impact of the disaster on the affected area
\end{tcolorbox}

\begin{figure}[h!]
  \centering
  \includegraphics[width=0.5\textwidth]{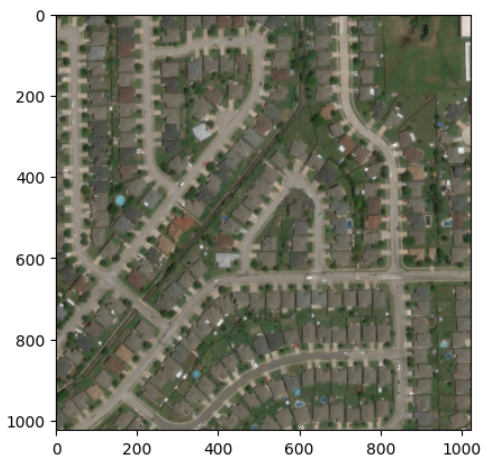}
  \caption{xBD without KG, QwenVL output.}
  \label{fig:example7b}
\end{figure}

\begin{tcolorbox}[colback=gray!5, colframe=black!50, coltitle=black, fonttitle=\bfseries, title=Baseline Caption (No KG) {[\ref{fig:example7b}]}]
\textbf{Caption:} the satellite image depicts an area that has been significantly impacted by tornado the visible damage is extensive with numerous buildings and structures \textcolor{purple}{\textbf{showing signs of destruction showing signs of destruction}} here is detailed description of the visible damage and its implications damage overview damaged areas buildings these areas show visible signs of damage such as showing signs of destruction \textcolor{purple}{\textbf{collapsed trees broken}} and debris scattered around the damage is concentrated in specific areas indicating that the tornado had localized impact undamaged \textcolor{purple}{\textbf{areasbuildings}} these areas appear to be relatively intact with no visible signs of damage from the tornado
\end{tcolorbox}

\subsection{Summary of Failure Modes}

Across all configurations, removing KG consistently produces four categories of errors: (1)~\textbf{hallucinations}- fabricated casualty counts and fictitious elements; (2)~\textbf{redundancy}- phrases repeated three or more times; (3)~\textbf{formatting errors}- concatenated words and missing spacing; and (4)~\textbf{semantic incoherence}- fragmented sentences. KG-enhanced models eliminate these failure modes across all architectures, VLMs, and datasets.
\end{appendices}

\bibliographystyle{IEEEtran}

\end{document}